\documentclass[journal]{IEEEtran}
\usepackage{amsmath,amsfonts}
\usepackage{algorithmic}
\usepackage{algorithm}
\usepackage{array}
\usepackage[caption=false,font=normalsize,labelfont=sf,textfont=sf]{subfig}
\usepackage{textcomp}
\usepackage{stfloats}
\usepackage{url}
\usepackage{verbatim}
\usepackage{graphicx}
\usepackage{cite}
\usepackage[pdftex, colorlinks,linkcolor={red},anchorcolor={blue},citecolor={green},urlcolor={magenta}, backref=page]{hyperref}

\usepackage{amsthm}

\usepackage{booktabs}
\usepackage{multirow}
\usepackage{pifont} 

\usepackage{bm}

\usepackage{color}
\usepackage{xcolor}
\definecolor{lightblue}{rgb}{0.12, 0.60, 0.92} 

\begin{document}

\title{Neuromorphic Object Detection: An In-Depth Study and Future Directions}

\author{Jianing~Li,~\IEEEmembership{Member,~IEEE,}
        Dianze~Li,
        Arren Glover,~\IEEEmembership{Member,~IEEE,}
        Xiaopeng~Fan,~\IEEEmembership{Senior~Member,~IEEE,}
        Guoqi~Li,
        Chiara~Bartolozzi,~\IEEEmembership{Member,~IEEE},
        Ryad~B.~Benosman,
        and~Yonghong~Tian,~\IEEEmembership{Fellow,~IEEE}

\thanks{Jianing Li is with the Beijing Key Laboratory of Brain-inspired Spiking Large Models, School of Computer Science, Peking University, Beijing 100871, China, and also with Peng Cheng Laboratory, Shenzhen 518000, China (e-mail: lijianing@pku.edu.cn).}
\thanks{Dianze Li is with the Beijing Key Laboratory of Brain-inspired Spiking Large Models, School of Computer Science, Peking University, Beijing 100871, China (e-mail: dianzeli@stu.pku.edu.cn).}
\thanks{Xiaopeng Fan is with the Department of Computer Science and Technology, Harbin Institute of Technology, Harbin, 150001, China, and also with Peng Cheng Laboratory, Shenzhen 518000, China (e-mail: fxp@hit.edu.cn).}
\thanks{Guoqi Li is with the Institute of Automation, Chinese Academy of Sciences, and Key Laboratory of Brain Cognition and Brain-inspired Intelligence Technology, Beijing 100045, China (e-mail: guoqi.li@ia.ac.cn).}
\thanks{Arren Glover and Chiara Bartolozzi are with the Event-Driven Perception for Robotics, Italian Institute of Technology, Genoa 16163, Italy (e-mail: arren.glover@iit.it and chiara.bartolozzi@iit.it).}
\thanks{Ryad B. Benosman is with the Robotics Institute, Carnegie Mellon University, Pittsburgh, PA 15213, USA, the University of Pittsburgh, Pittsburgh, PA 15260, USA, and also with the Sorbonne Université, INSERM, CNRS, Institut de la Vision, 75012 Paris, France (e-mail: benjry.benos@gmail.com).}
\thanks{Yonghong Tian is with the Beijing Key Laboratory of Brain-inspired Spiking Large Models, School of Computer Science, Peking University, Beijing 100871, China, with the School of AI for Science, Shenzhen Graduate School, Peking University, Shenzhen 518055, China, and also with Peng Cheng Laboratory, Shenzhen 518000, China (e-mail: yhtian@pku.edu.cn).}
\thanks{Manuscript received April 23, 2025; revised November 29, 2025; revised May 19, 2026; revised July 10, 2026; accept July 22, 2026.}
\thanks{The first two authors contributed equally.}
\thanks{The corresponding author: Yonghong Tian.}
}

\markboth{PROCEEDINGS OF THE IEEE}
{Shell \MakeLowercase{\textit{et al.}}: A Sample Article Using IEEEtran.cls for IEEE Journals}


\maketitle

\begin{abstract}
Conventional frame-based cameras face significant challenges in detecting objects under high-speed motion blur or in low-light environments. Neuromorphic cameras provide asynchronous visual streams with high temporal resolution and a wide dynamic range, offering a promising solution for object detection under challenging conditions. Despite the development of numerous models and the emergence of various applications in neuromorphic object detection, there is still a lack of deep understanding and standardized benchmarks to assess progress and address key challenges. In this paper, we provide a comprehensive survey and benchmark of existing neuromorphic object detection algorithms. Specifically, we first present a problem description, review the available datasets, and revisit the evaluation metrics. We then explore existing neuromorphic object detection approaches from various perspectives, including event representation, temporal modeling, multimodal fusion, asynchronous processing, low-latency processing, and energy-efficient computing. Furthermore, we evaluate a wide range of representative neuromorphic object detection models and offer detailed analyses of the comparative results. Finally, we discuss unresolved issues in neuromorphic object detection and propose potential future research directions. We hope this survey and benchmark will be a valuable resource for researchers and provide guidance for future advancements in neuromorphic object detection. 
\end{abstract}


\section{Introduction}
\IEEEPARstart{O}{bject} detection~\cite{liu2020deep, zou2023object, cheng2023towards} is a fundamental and longstanding topic with a wide range of applications, such as agile robotics~\cite{martinez2017object, du2018unmanned}, autonomous driving~\cite{mao20233d},  surveillance~\cite{nascimento2006performance}, and augmented reality~\cite{wang2022leaf+}. The classical visual paradigm focuses on designing powerful algorithms for detecting objects in conventional frames. Nevertheless, frame-based object detection performance may significantly drop in fast motion blur~\cite{sayed2021improved} or challenging light scenes~\cite{onzon2021neural}. Beyond these issues, the frame-based paradigm also suffers from high data redundancy and increased power consumption during high-speed capture, making it difficult to deploy on resource-constrained edge devices under such conditions. Consequently, researchers are motivated to explore novel sensing paradigms that overcome the limitations of conventional cameras for object detection tasks in challenging scenarios.

\subsection{What Is Neuromorphic Object Detection About?}
Neuromorphic cameras (e.g., DVS~\cite{lichtsteiner2006128, finateu20201280x720}, ATIS~\cite{posch2010qvga}, DAVIS~\cite{brandli2014240}, and CeleX~\cite{guo2016live}) capture visual information using asynchronous event streams instead of fixed frames. This novel sensing paradigm offers some significant advantages over conventional frame-based cameras, including high temporal resolution and low latency (both in the order of microseconds), high dynamic range (120 dB versus 60 dB of standard cameras), low power consumption, and reduced redundancy. As a result, neuromorphic cameras are becoming more widely utilized in various computer vision tasks (e.g., video reconstruction~\cite{zhu2022event}, depth estimation~\cite{liu2024event, liu2025high}, and object tracking~\cite{wang2023visevent}) and agile robotics tasks (e.g., obstacle avoidance~\cite{falanga2020dynamic} and drone landing~\cite{paredes2024fully}).

\begin{figure}[t]
\centering
\includegraphics[width=\linewidth]{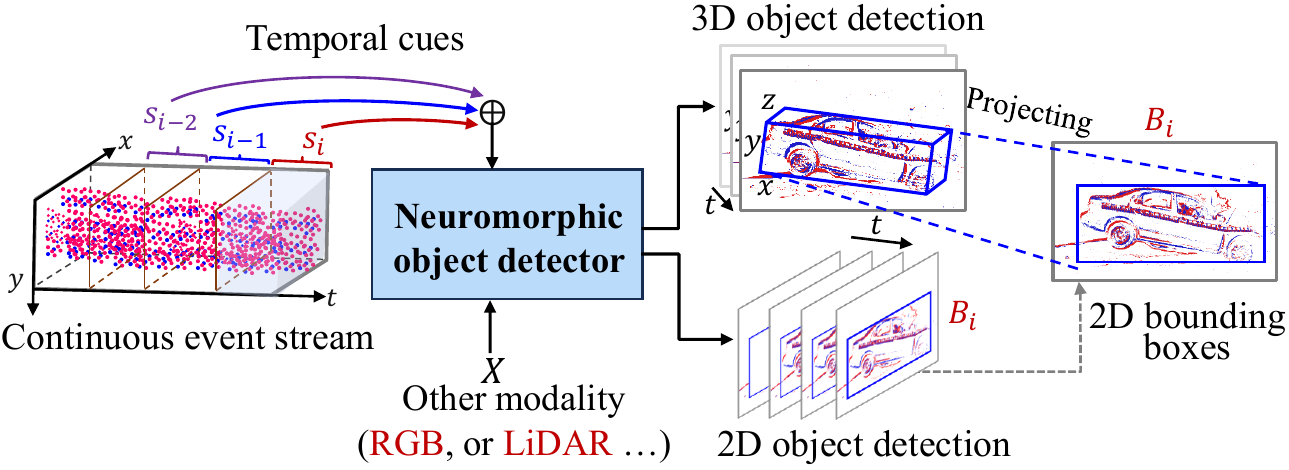}
\caption{An illustration of neuromorphic object detection. The continuous event stream is usually first divided into discrete temporal event bins as the basic processing unit. The designed neuromorphic object detector can leverage rich temporal cues from event bins for 2D or 3D object detection. In specific scenes, incorporating other modalities (e.g., RGB frames~\cite{li2023sodformer, gehrig2024low} or LiDAR point clouds~\cite{li2021enhancing}) improves object detection accuracy.}
\label{fig:problem_definition}
\end{figure}

``Neuromorphic object detection" or ``object detection using neuromorphic cameras" is a novel research field that combines neuromorphic computing and computer vision~\cite{li2022asynchronous, li2022retinomorphic, li2023sodformer}. Its primary goal is to detect spatiotemporal locations and identify categories of specific objects using neuromorphic cameras (see Fig.~\ref{fig:problem_definition}). Neuromorphic object detectors are generally evaluated using the following criteria~\cite{liu2020deep, su2023deep}: (i) \emph{Accuracy} - Achieving high accuracy in detecting and classifying objects; (ii) \emph{Computational efficiency} - Enabling high inference speed within low memory usage; (iii) \emph{Power consumption} - Implementing energy-efficient models with low power consumption;  (iv) \emph{Robustness} - Exhibiting consistent performance across variations in motion speed, light intensity, and other factors.

\begin{figure*}[htbp]
\centering
\centerline{\includegraphics[width=\linewidth]{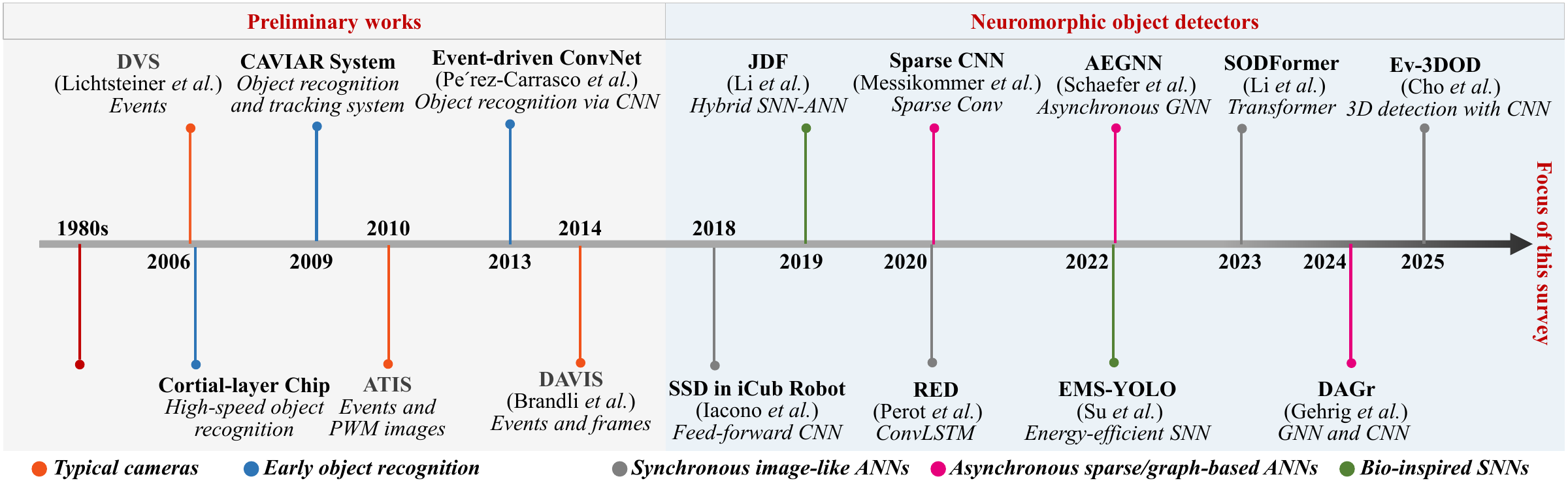}}
\vspace{-0.20cm}
\caption{Timeline of key advances and representative milestones in neuromorphic object detection. Preliminary efforts include the concept of neuromorphic~\cite{Mead1989, mead1990neuromorphic}, the development of typical event cameras (e.g., DVS~\cite{lichtsteiner2006128}, ATIS~\cite{posch2010qvga}, and DAVIS~\cite{brandli2014240}), and the early neuromorphic systems capable of high-speed object recognition (e.g., cortical-layer processing chip~\cite{serrano2006neuromorphic}, CAVIAR~\cite{serrano2009caviar}, and event-driven ConvNet~\cite{perez2013mapping}). Neuromorphic object detectors can be broadly grouped into three processing paradigms: synchronous image-like ANNs, asynchronous sparse/graph-based ANNs, and bio-inspired SNNs. The first category includes feed-forward CNNs (e.g., SSD on the iCub robot~\cite{iacono2018towards} and AED~\cite{liu2023motion}), temporally-aware RNNs (e.g., RED with ConvLSTM~\cite{perot2020learning} and HMNet with latent memory~\cite{hamaguchi2023hierarchical}), and transformer-based models (e.g., SODFormer~\cite{li2023sodformer} and S5-ViT with SSM~\cite{zubic2024state}). The second category primarily consists of sparse convolution models (e.g., Sparse CNN~\cite{messikommer2020event}) and graph-based approaches (e.g., AEGNN~\cite{schaefer2022aegnn}, DAGr~\cite{gehrig2024low}, and EGSST~\cite{wu2024egsst}). The third category mainly contains pure SNNs (e.g., EMS-YOLO~\cite{su2023deep} and EAS-SNN~\cite{wang2024eas}) and hybrid SNN-ANN models (e.g., JDF~\cite{li2019event} and HDI-Former~\cite{li2026rethinking}). More recently, the field has also expanded toward 3D object detection (e.g., Ev-3DOD~\cite{cho2025ev}). Most of the listed works in the chronology are foundational or widely recognized contributions.}
\label{fig:milestones}
\vspace{-0.30cm}
\end{figure*}

\subsection{History of Neuromorphic Object Detection}
Neuromorphic object detection is an emerging frontier and a fundamental research topic in the neuromorphic community~\cite{gallego2020event, li2021recent}. Early and representative works (e.g., cortical-layer processing chip~\cite{serrano2006neuromorphic}, CAVIAR~\cite{serrano2009caviar}, and event-driven ConvNet~\cite{perez2013mapping}) design neuromorphic systems capable of high-speed object recognition. For example, a neuromorphic cortical-layer microchip~\cite{serrano2006neuromorphic} for event-based vision systems demonstrates the ability to discriminate between rotating propellers of different shapes at speeds of up to 5,000 revolutions per second. The CAVIAR project~\cite{serrano2009caviar} implements a massively parallel neuromorphic processing system that achieves object recognition and tracking with millisecond latency. Similarly, event-driven convolutional neural networks (ConvNet)~\cite{perez2013mapping} are designed to recognize rotating human silhouettes or high-speed poker card symbols. Research in neuromorphic object detection takes shape alongside the development of typical event cameras (e.g., DVS~\cite{lichtsteiner2006128}, ATIS~\cite{posch2010qvga}, and DAVIS~\cite{brandli2014240}). Existing neuromorphic object detectors can be broadly categorized based on two orthogonal properties of data representation and activation functions in neural networks. This yields four conceptual quadrants: synchronous image-like artificial neural network (ANN) models~\cite{liu2016combined, perot2020learning, gehrig2023recurrent, li2023sodformer}, asynchronous sparse/graph-based ANNs~\cite{messikommer2020event, schaefer2022aegnn, gehrig2024low}, bio-inspired spiking neural networks (SNNs)~\cite{cordone2022object, su2023deep}, and binary-activation ANNs~\cite{cladera2020device}. Among these, the first three categories have been actively explored in neuromorphic object detection, whereas the fourth remains largely underexplored. As summarized in Fig.~\ref{fig:milestones}, most of the works listed in the chronology represent foundational or widely recognized contributions to the topic. Therefore, the main technical discussion is structured around these three predominant categories as follows.

Synchronous image-like ANNs~\cite{hu2020learning, perot2020learning} typically convert event streams into image-like representations and apply standard ANN-based object detectors. This paradigm is widely used and remains the mainstream approach for neuromorphic object detection. Early works~\cite{liu2016combined, iacono2018towards, li2019event, jiang2019mixed} usually convert continuous events into 2D image-like representations (e.g., event images) and then utilize feed-forward frame-based models to detect objects. While these straightforward strategies directly shift frame-based models to the event domain, they are hard to make the best of the spatiotemporal attributes of event streams. After Perot~\emph{et al.}~\cite{perot2020learning} released two temporally long-term datasets (i.e., Gen1 Detection and 1Mpx Detection) and presented a recurrent architecture (i.e., RED), subsequent works attempt to design RNN models~\cite{li2022asynchronous, wang2023dual, andersen2022event, silva2024recurrent, zhu2024spatio}, temporal aggregations~\cite{zhou2023rgb, cao2023chasing, wu2024leod}, or temporal transformers (e.g., SODFormr~\cite{li2023sodformer}, RVT-B~\cite{gehrig2023recurrent}, and S5-ViT with SSM~\cite{zubic2024state}) to leverage rich temporal cues from continuous event streams. Nevertheless, these temporally-aware models improve accuracy while introducing higher computational complexity. More recently, the field has also expanded toward 3D neuromorphic object detection (e.g., Ev-3DOD~\cite{cho2025ev} and Ev-Stereo3D~\cite{kang2025unleashing}).

Asynchronous sparse/graph-based ANNs~\cite{messikommer2020event, schaefer2022aegnn, gehrig2022pushing, gehrig2024low} operate directly on event streams in an event-by-event or asynchronous manner, making them a promising paradigm for low-latency object detection. In contrast to synchronous methods, the asynchronous event-based paradigm handles each event individually instead of processing image-like representations. For example, sparse convolution~\cite{messikommer2020event} efficiently handles the asynchronous nature of event data. Graph neural networks (GNNs)~\cite{schaefer2022aegnn, gehrig2022pushing} provide a powerful tool for modeling the spatiotemporal relationships for sparse events. Although graph-based object detectors (e.g., AEGNN~\cite{schaefer2022aegnn} and EGSST~\cite{wu2024egsst}) achieve low latency through event-by-event processing, their detection accuracy on benchmark datasets (e.g., Gen1 Detection~\cite{de2020large}) often remains below that of state-of-the-art synchronous image-based models (e.g., S5-ViT~\cite{zubic2024state} and EvRT-DETR-B~\cite{torbunov2025evrt}). A promising approach is hybrid object detectors (e.g., AGr~\cite{gehrig2024low} and ACGR~\cite{li2025asynchronous}) that combine a standard CNN for image processing with an efficient asynchronous GNN for event data. Such hybrid frameworks achieve both high accuracy and low-latency object detection on the DSEC-Detection dataset~\cite{gehrig2024low}.

Bio-inspired SNNs~\cite{zenke2021brain, fang2023spikingjelly, eshraghian2023training} are well-suited for energy-efficient object detection by processing asynchronous events via spike-driven neuron dynamics with binary activations. Early works are mainly unsupervised SNN-based methods~\cite{nagaraj2023dotie, bulzomi2023object}, hybrid SNN-ANN frameworks~\cite{li2019event, kugele2021hybrid, li2026rethinking}, and ANN-to-SNN models~\cite{wang2023spike}. With the introduction of alternative gradients into deep SNNs, there is an increasing exploration of directly-trained SNNs~\cite{cordone2022object, su2023deep, yuan2023trainable, fan2024sfod, luo2024integer} for object detection. These models achieve higher accuracy with fewer simulation steps than conversion-based SNNs on event-based datasets (e.g., Gen1 Detection). Despite their energy efficiency advantages over conventional ANNs, most existing SNN-based object detectors still operate synchronously on image-like event representations rather than leveraging fully event-driven computation during training. The main reason is that most existing deep SNN training strategies remain tightly coupled to timestep-based computation on GPUs. Although event-driven SNNs that process raw event streams asynchronously on an event-by-event basis have begun to be explored for classification tasks (e.g., ASGCN~\cite{zeng2025leveraging}), they have been applied only rarely to complex regression tasks like object detection. Therefore, developing event-driven SNNs for object detection remains an open challenge,  particularly in demonstrating verified low-latency and low-energy performance on neuromorphic computing chips~\cite{merolla2014million, davies2021advancing, pei2019towards, shen2016darwin}.

Neuromorphic object detection has emerged as a rapidly advancing research topic. While existing surveys cover event-based vision~\cite{gallego2020event, chen2020event, iddrisu2024event, ghosh2025event}, SNNs~\cite{eshraghian2023training, tavanaei2019deep, rathi2023exploring, deng2020rethinking}, neuromorphic hardware~\cite{shrestha2022survey, li2021recent, sandamirskaya2022neuromorphic, posch2014retinomorphic, chen2011pulse}, and neuromorphic computing~\cite{schuman2022opportunities, indiveri2015memory, li2024brain, roy2019towards, yik2025neurobench}, there is a lack of focus on the topic of neuromorphic object detection. As a result, it is necessary to review existing models, benchmark algorithms, and provide novel insight for ongoing progress, especially for newcomers entering the neuromorphic community.

\subsection{Our Contributions}
The goal of this paper is to present an in-depth survey and benchmark of existing neuromorphic object detectors\footnote[1]{\url{https://github.com/jianing-li/awesome-neuromorphic-object-detection}}. Technically, we begin by describing neuromorphic sensing mechanisms, presenting the problem description, summarizing datasets, and revisiting evaluation metrics. Then, we review neuromorphic object detection algorithms considering event representation, temporal modeling, multimodal fusion, asynchronous processing, low-latency processing, and energy-efficient computing. Moreover, we evaluate representative object detection models and offer in-depth analyses of comparison results. Finally, we discuss open challenges and highlight promising future directions in neuromorphic object detection. The main contributions of this work are summarized as:
\begin{itemize}
\item We present a \emph{\textbf{comprehensive survey}} of problem description, datasets, evaluation metrics, and neuromorphic object detectors from various perspectives. This in-depth survey aims to assist newcomers in gaining a deeper understanding of the key features of the field.
\item We conduct \emph{\textbf{extensive and systematic experiments}} to quantitatively compare state-of-the-art neuromorphic object detection models. Our benchmarks may guide readers to easily grasp the similarities and differences among various neuromorphic object detectors.
\item We give an overview of \emph{\textbf{open issues and future directions}} in the field of neuromorphic object detection. These open problems may be valuable for researchers and serve as guidance for future innovations in the community.
\end{itemize}

The rest of this paper is organized as follows. Section~\ref{sec:mechanism} reviews neuromorphic sensing principles. Section~\ref{sec:problem_challenge} presents the problem statement and main challenges. Sections~\ref{sec:dataset} and~\ref{sec:metric} describe datasets and evaluation metrics. Section~\ref{sec:method} reviews neuromorphic object detection algorithms, and Section~\ref{sec:experiment} presents benchmarking results. Finally, Sections~\ref{sec:discussion} and~\ref{conclusion} highlight future research directions and draw conclusions.

\section{Neuromorphic Sensing Preliminaries}\label{sec:mechanism}
This section first presents the sensing mechanisms of two types of neuromorphic cameras and then analyzes how their working principles relate to the object detection task.

\subsection{Dynamic Visual Sensing}\label{sec:dynamic_sensing}
Dynamic vision sensors (DVS)~\cite{lichtsteiner2008128, posch2014retinomorphic, gallego2020event}, also known as event cameras, emulate the retina’s dynamic information processing by abstracting the photoreceptor–bipolar–ganglion cell pathway. In contrast to conventional frame-based cameras, each pixel in DVS responds to changes in light intensity $I(x,y,t)$ by generating a stream of events. Specifically, an event $e_{n}$ is a four-attribute tuple $(x_{n}, y_{n}, t_{n}, p_{n})$ using the addressing event representation (AER)~\cite{sivilotti1991wiring, boahen1996retinomorphic, boahen2000point}. It is triggered for the pixel $(x_{n}, y_{n})$ at the timestamp $t_{n}$ when the log-intensity changes beyond the pre-defined threshold $\theta_{1}$. The dynamic visual sensing mechanism can be depicted as follows:
\begin{eqnarray}
\text{ln} I(x_{n}, y_{n}, t_{n})-\text{ln} I(x_{n}, y_{n}, t_{n}-\Delta t_{n}) =p_{n} \theta_{1},
\end{eqnarray}
where the polarity $p_{n}$$\in$$\left\lbrace 1, -1 \right\rbrace $ indicates whether the brightness is increasing or decreasing, and $\Delta t_{n}$ is the time elapsed since the last event at the same pixel.

Notably, this asynchronous pixel-by-pixel encoding with AER fundamentally differs from synchronous frames. Consequently, such event streams, appearing as sparse and discrete points in the spatiotemporal domain~\cite{fu2019spike, gu2021spatio, xiang2022temporal, wang2026event}, remain incompatible with direct processing by frame-based object detectors. Intuitively, this event stream can be described as:

\begin{eqnarray}
S\left(x,y,p, t\right)=\left\lbrace  p_{n}\delta\left(x-x_{n},y-y_{n},t-t_{n}\right)    \right\rbrace _{n=1}^{N_{e}},
\end{eqnarray}
where $N_{e}$ is the number of event data during a spatiotemporal window, and $\delta\left(\cdot\right)$ refers to the Dirac delta function, with $\delta\left(t\right)=0, \forall$$t\neq0$, and $\int$$\delta\left(t\right)dt=1$. 

In general, existing event cameras can be roughly divided into three categories. The first category is the raw DVS that outputs only dynamic events, and its main types are DVS128~\cite{lichtsteiner2008128} from IniVation, EKV4~\cite{finateu20201280x720} from Prophesee, and DVS-Gen4~\cite{suh20201280} from Samsung, etc. The second category is an asynchronous time-based image sensor (ATIS)~\cite{posch2010qvga} that contains a DVS circuit that triggers another circuit to measure absolute intensity. Similarly, each unsigned event (i.e., without polarity) is triggered in CeleX-V~\cite{chen2019live} from CelePixel, and the absolute light intensity for that pixel is measured. Consequently, an intensity frame is stored in an external memory buffer, and the absolute light of each pixel is updated for each new event. The third category is dynamic and active pixel vision sensor (DAVIS) that output dynamic events and frames in the same pixel array, and its main types are DAVIS240~\cite{brandli2014240} and DAVIS346~\cite{moeys2017sensitive} from IniVation, OV60B10~\cite{guo2023three, yu2024learning} from OmniVision, Sony’s hybrid vision sensors~\cite{niwa20232, kodama20231}, ALPIX~\cite{yunfan2024rgb} from AlpsenTek, and Tianmouc~\cite{yang2024vision} from Tsinghua University, etc. In ATIS or DAVIS event cameras, the term ``DVS" often refers to the event output mode.

\subsection{Pulse-modulation Imaging}\label{sec:integrating_sensing}
Time-based image sensors~\cite{chen2011pulse}, namely pulse-modulation image sensors, mimic the visual sampling mechanism of the central fovea in the retina~\cite{huang20231000}. These bio-inspired cameras utilize the spike frequency or inter-spike interval to encode the light intensity for each independent pixel. A one-bit spike $(x_{n}, y_{n}, t_{n})$ without the polarity is fired when the accumulator of light intensity $I(x,y,t)$ exceeds the pre-defined threshold $\theta_{2}$. Pulse-modulation imaging is also known as pulse-width modulation (PWM) encoding or pulse frequency modulation (PFM) encoding~\cite{culurciello2003biomorphic, xu2020denoising}, which can be formulated as:
\begin{eqnarray}
\int_{t_{n-1}}^{t_{n}} I(x_{n},y_{n}, t) dt = \theta_{2},
\end{eqnarray}
where $t_{n-1}$ is the timestamp of the previous spike at the same pixel. The inter-spike interval $\Delta t_{n}$$=$$(t_{n} - t_{n-1})$ exhibits an inverse relationship with light intensity, resulting in higher firing rates under brighter illumination conditions. When a spike occurs, it is recorded as ``1" at the corresponding pixel; otherwise ``0" is recorded. Therefore, spatiotemporal spikes can be presented as follows:
\begin{eqnarray}
S\left(x,y,t\right)=\left\lbrace \delta\left(x-x_{n},y-y_{n},t-t_{n}\right)    \right\rbrace _{n=1}^{N_{s}},
\end{eqnarray}
where $N_{s}$ s the number of spatiotemporal spikes.

Pulse-modulation imaging~\cite{chen2011pulse} has a long development history within the neuromorphic sensing community. An early and representative work~\cite{culurciello2003biomorphic} designs an arbitrated address-event image sensor that uses pulse-frequency modulation to encode light intensity into asynchronous spikes. This is followed by a series of time-based image sensors employing asynchronous event-based readout, including time-to-first-spike (TTFS) imager sensor~\cite{guo2007time}, time-to-first-spike (TFS) image sensors~\cite{shoushun2007arbitrated}, and ATIS~\cite{posch2010qvga}. Other designs have implemented synchronous frame-based readout, such as the spiking-pixel image sensor~\cite{shoushun2008robust} and the Vidar Gen1 sensor~\cite{dong2017spike, huang20231000}. Notably, the latter utilizes a high-speed synchronous scanned readout that increases bandwidth requirements when the pixel array generates sparse spikes, distinguishing it from asynchronous event-based architectures. It is worth noting that pulse-modulation imaging encodes light intensity as one-bit spikes rather than typical 8-bit frames. Thus, such spatiotemporal spike streams may not be directly suited to existing standard frame-based object detectors.

\subsection{Comparison with Two Sensing Mechanisms}

From the sensing principles of the two neuromorphic camera types, event cameras encode light intensity changes as asynchronous events, whereas time-based image sensors convert light intensity into one-bit spikes. As a result, event cameras exhibit lower data redundancy and higher temporal resolution, while the frame-free pulse-modulation imaging paradigm~\cite{chen2011pulse} enables time-based image sensors (e.g., Vidar Gen1) to reconstruct finer-textured frames~\cite{zhu2020retina, xiang2021learning, zhu2022ultra, zhu2021neuspike} even under fast motion scenarios.

In particular, these two visual sensing schemes produce output formats that fundamentally differ from synchronous frames. While this makes them particularly suitable for object detection in high-speed motion scenarios, it also means they cannot be directly applied to standard frame-based object detectors. For instance, event cameras serve as natural motion detectors and are well-suited for object detection in challenging scenarios. While single-modality tasks~\cite{perot2020learning, li2022asynchronous} may achieve satisfactory results by only processing events, they often struggle to capture static textures and achieve high-precision detection. Consequently, several joint frameworks~\cite{li2019event, li2023sodformer} have been proposed to complement event streams with other modalities (e.g., RGB frames~\cite{li2023sodformer, gehrig2024low} or LiDAR point clouds~\cite{li2021enhancing}), thereby improving overall accuracy and robustness. In contrast, time-based image sensors offer a simple yet effective way to reconstruct the image before performing object detection~\cite{li2022retinomorphic}, although their applications in object detection remain relatively underexplored. Driven by the rapid advances of modern DVS families (e.g., DAVIS346~\cite{moeys2017sensitive}, EVK4~\cite{finateu20201280x720}, and CeleX-V~\cite{chen2019live}), event cameras have become the dominant sensing platform in the neuromorphic vision community. In addition, the expanding ecosystem of open-source datasets (e.g., DSEC~\cite{gehrig2021dsec} and PKU-DAVIS-SOD~\cite{li2023sodformer}), community-maintained resources\footnote[2]{\url{https://github.com/uzh-rpg/event-based_vision_resources}}, and software tools (e.g., ESIM~\cite{rebecq2018esim}, V2E~\cite{hu2021v2e}, and SpikingJelly~\cite{fang2023spikingjelly}) have further accelerated progress in the field. For these reasons, the subsequent sections of this paper focus primarily on neuromorphic object detection using event cameras.

\section{Task Description and Challenges} \label{sec:problem_challenge}
In this section, we first present a problem description of neuromorphic object detection and then summarize the main challenges in the research community.

\subsection{Problem Statement} \label{sec:definition}
``Neuromorphic object detection" aims to obtain spatiotemporal locations and identify categories of specific objects in visual streams captured by neuromorphic cameras. In practice, most existing neuromorphic object detection frameworks follow a similar processing pipeline. First, event bins are extracted from the continuous event stream using either fixed or adaptive time intervals or through activity-based grouping. These event bins serve as fundamental processing units compatible with both synchronous and asynchronous processing schemes. Second, the event bins are processed through one of two primary paradigms: (i) synchronous frame-based methods that convert events into 2D image-like representations for detection models, or (ii) asynchronous event-driven approaches that process events using event-by-event architectures, such as GNN-based models or event-driven SNNs. It is important to note that neuromorphic object detection typically requires extracting stable features over short temporal windows before generating bounding boxes, as reliably deriving object contours from individual events is generally infeasible. As illustrated in Fig.~\ref{fig:problem_definition}, this provides an intuitive visual example of the entire pipeline from event stream processing to bounding box prediction. To quickly gain a formal understanding of this topic, we present the problem formulation of neuromorphic object detection as follows.

Let asynchronous event stream $S(x,y,p,t)$ be divided into $K$ event temporal bins $S$=$\left\lbrace S_{1}, S_{2}, ..., S_{K} \right\rbrace$. For the current timestamp $t_{i}$, the object spatiotemporal location and category information (i.e., bounding boxes) can be computed by leveraging temporal cues from multiple adjacent event bins $\left\lbrace S_{i-k}, ..., S_{i} \right\rbrace$, $k\in[0, K]$. This function $\mathcal{D}$ is called neuromorphic object detection as:
\begin{eqnarray}
    B_{i}=\mathcal{D}\left( S_{i-k}, ..., S_{i} \right), \label{eq:problem_formulation}
\end{eqnarray}
where $B_{i}$ is the predicted result of 2D object detection~\cite{li2021asynchronous, li2023sodformer} or 3D object detection~\cite{cho2025ev} in the timestamp $t_{i}$. $S_{i}= \left\lbrace e_{n} \right\rbrace _{n=1}^{N_{e}}$ represents the spatiotemporal events within an event bin. The parameter $k$ determines the length of leveraging temporal cues from adjacent event bins.

\begin{figure}[t]
\centering
\includegraphics[width=\linewidth]{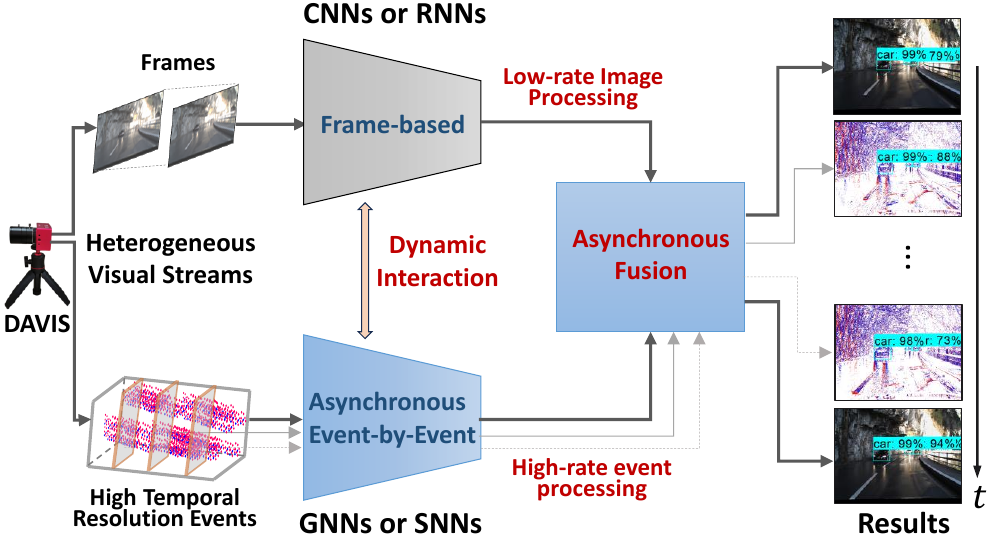}
\caption{A representative pipeline of neuromorphic object detection using frames and events. The framework integrates two visual streams from a DAVIS camera to achieve high-rate object detection in an asynchronous processing manner. It leverages the complementary advantages of each modality through dynamic interaction between a frame-based ANN operating on low-rate images and an asynchronous GNN or event-driven SNN that processes events.}
\label{fig:asynchronous_processing}
\end{figure}

In general, achieving high-accuracy object detection using only DVS events can be challenging in certain scenarios, such as slow-motion or static scenes. Consequently, the framework can be flexibly extended to multimodal object detection by incorporating additional sensing modalities, such as RGB frames~\cite{li2023sodformer, gehrig2024low}, infrared frames~\cite{hu2020learning}, or LiDAR point clouds~\cite{li2021enhancing}. As shown in Fig.~\ref{fig:asynchronous_processing}, we present a representative pipeline for neuromorphic object detection that integrates events and frames in an asynchronous processing manner. In short, neuromorphic object detection is not limited to single-modal models, it can also be more broadly extended to multimodal methods that incorporate neuromorphic data.

Note that, neuromorphic object detection operates differently from conventional frame-based methods (see Eq.~\ref{eq:problem_formulation}). It possesses \emph{\textbf{three unique properties}}, precisely reflecting the term ``neuromorphic": (i) Inputs are asynchronous events instead of frames; (ii) Accurate object detection can be achieved by using rich temporal cues rather than each single image; (iii) Continuous event streams with high temporal resolution enable object detection at any time, overcoming limitations in inference frequency from conventional frame-based cameras.

\subsection{Main Challenges}\label{sec:challenge}
\begin{figure}[t]
\centering
\includegraphics[width=0.69\linewidth]{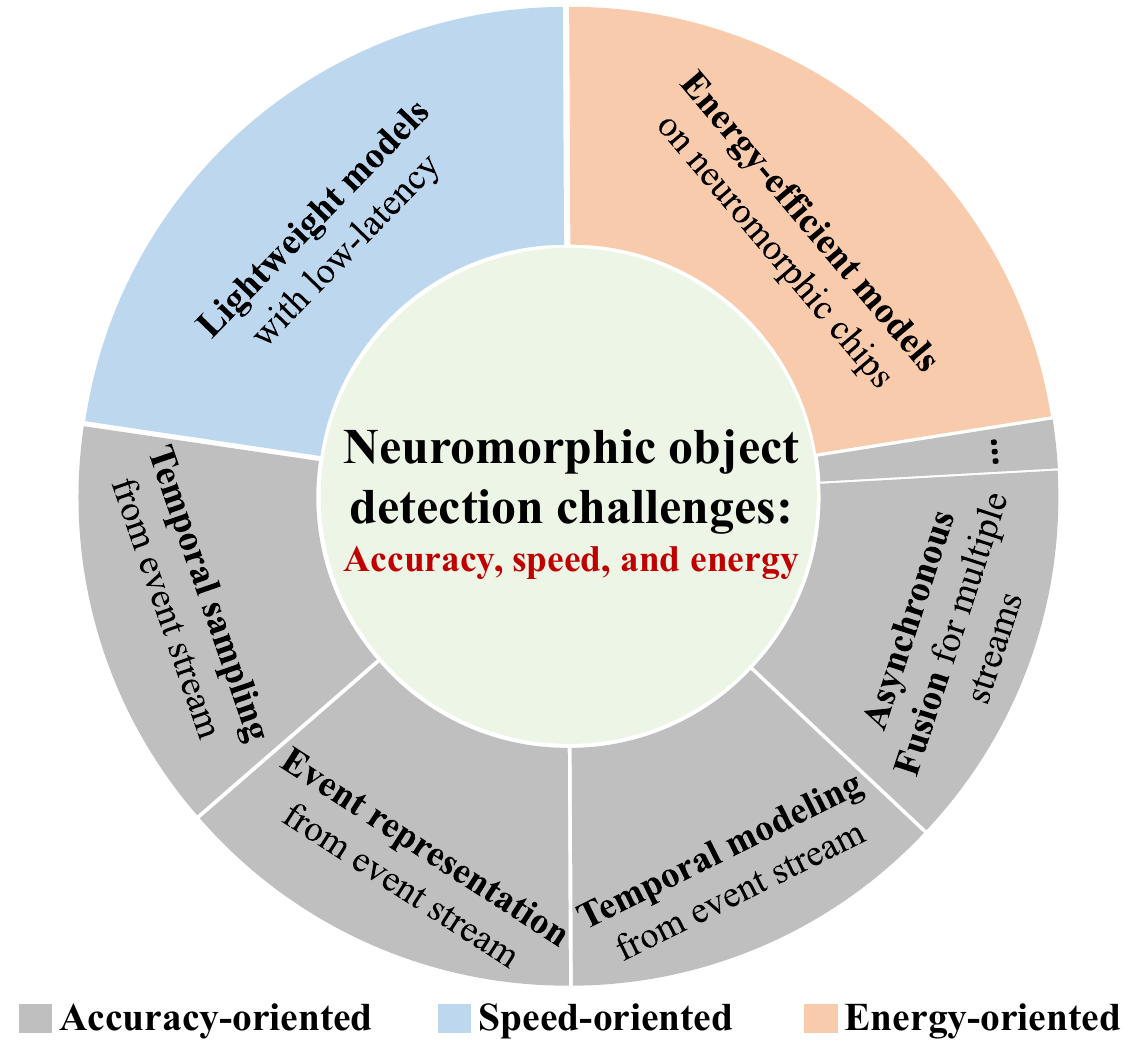}
\caption{Main challenges of neuromorphic object detection. Neuromorphic cameras operate differently from conventional frame-based sensors, posing challenges in three key aspects: detection accuracy, computational speed, and energy efficiency. Accuracy-oriented challenges focus on designing effective modules, such as temporal sampling, event representation, temporal modeling, and asynchronous fusion. Speed-oriented challenges involve developing lightweight models capable of fast inference. Energy-oriented challenges lie in designing highly energy-efficient models. In practice, colors highlight each factor’s primary impact, while some factors overlap across aspects.}
\label{fig:neuromorphic_detection_challenges}
\end{figure}

Neuromorphic object detection seeks to design a general algorithm capable of excelling in detection accuracy, computational efficiency, and power consumption. Nevertheless, the sparse nature of asynchronous events makes them incompatible with existing frame-based algorithms and poses significant challenges in achieving the stated objectives (see Fig.~\ref{fig:neuromorphic_detection_challenges}). Firstly, the accuracy-oriented challenges focus on designing effective modules, such as temporal sampling, event representation, temporal modeling, and asynchronous fusion. Secondly, the speed-oriented challenges involve developing lightweight or event-driven models capable of fast inference, particularly on resource-constrained devices. Thirdly, the energy-oriented challenges lie in designing highly energy-efficient models, especially for deployment on neuromorphic chips. A more in-depth analysis of each challenge is presented as follows.

\textbf{Temporal Sampling Challenges.} The continuous event stream typically needs to be divided into discrete temporal event bins as the basic preprocessing units. Most existing temporal sampling strategies directly split event streams with a constant temporal window $\Delta t$ or a constant event number $N$. However, this simple sampling may result in temporal bins with textures that are either too sparse in fast motion or excessively dense in slow motion~\cite{li2022asynchronous, fischer2020event}. In other words, these fixed slicing strategies might neglect the temporal information embedded in the event stream, limiting their ability to generalize event bins across various motion scenarios. Ideally, an effective event bin trigger should dynamically determine when to segment the stream without relying on manual hyperparameter tuning. A representative example is STARE~\cite{chen2026bridging}, which presents a novel paradigm for dynamic event stream sampling and effectively eliminates the need for pre-sliced bins in latency-aware perception systems. Overall, designing an adaptive temporal sampling strategy that generates effective event bins robust to varying motion speeds remains a significant challenge.

\textbf{Event Representation Challenges.} Discrete event bins are usually transformed into 2D image-like representations, directly compatible with mainstream frame-based object detectors~\cite{li2022asynchronous, zubic2023chaos}. Early attempts mainly design a kernel function to map asynchronous events into handcrafted representations (e.g., event images~\cite{maqueda2018event}, voxel grids~\cite{zhu2019unsupervised}, and time surfaces~\cite{lagorce2016hots}). The latest efforts have been made in designing end-to-end learning representations (e.g., EST~\cite{gehrig2019end}, Matrix-LSTM~\cite{messikommer2020event}, and event embeddings~\cite{tulyakov2019learning}) via ANNs or SNNs. Here, event embeddings refer to learned vector representations that map raw asynchronous events into a dense feature space. These event representations can directly consume frame-based object detectors and achieve state-of-the-art performance. In fact, how to design a powerful yet efficient event representation that leverages spatiotemporal attributes to maximize object detection performance remains an open issue.

\textbf{Temporal Modeling Challenges.} Continuous event streams provide rich temporal cues along the temporal dimension, capturing dynamic spatiotemporal patterns such as object motion and appearance changes across timestamps. Most existing event-based object detectors typically utilize a feed-forward method to achieve satisfactory performance, where they independently run feed-forward frame-based detection models on each image-like representation. Yet, object detection using an isolated image-like representation may encounter difficulties in specific scenarios, including weak texture, occlusion, or out-of-focus issues. For example, a single snapshot may capture only a partially visible object due to occlusion, making detection unreliable. By leveraging temporal cues, the detector can locate the object across timestamps, using motion trajectories and features observed before and after the occluded region. Similarly, when an object is blurred in out-of-focus regions, temporal cues in the event stream can capture sharp edges or motion patterns as the object moves between positions. Overall, the continuous event stream contains rich temporal information that allows the detector to identify and locate similar objects across time. Thus, effectively modeling temporal correspondence from the continuous event stream for object detection still remains an ongoing challenge.

\textbf{Asynchronous Fusion Challenges.} Leveraging events in conjunction with other modalities to enhance object detection accuracy has become increasingly popular~\cite{hu2020learning, li2022retinomorphic}. Continuous event stream poses a challenge for existing frame-based multimodal frameworks in managing multiple heterogeneous sensing streams. Most existing fusion methods segment event streams into discrete event bins to match the frequencies of other modalities, and then integrate synchronized streams through post-processing or feature aggregation operations. Some works~\cite{li2023sodformer, gehrig2021combining} also explore asynchronous fusion strategies aimed at enhancing the frequency of task inference. So far, how to design asynchronous multimodal fusion strategies for multiple heterogeneous sensing streams in the object detection task remains an unresolved challenge.

\textbf{Low-latency Processing Challenges.} Event cameras provide microsecond-level time resolution, making them particularly well-suited for high-speed object detection~\cite{mitrokhin2019learning, falanga2020dynamic}. Nowadays, most neuromorphic object detection methods~\cite{perot2020learning, li2023sodformer} primarily focus on maximizing accuracy by designing increasingly complex modules. Yet, these methods may result in slower inference speeds, potentially deviating from the initial goal of designing event cameras for high-speed sensing. A minority of studies explore lightweight frameworks~\cite{peng2023better} or asynchronous graph-based processing models~\cite{schaefer2022aegnn, gehrig2022pushing} to reduce inference speed. Thus, designing lightweight neuromorphic object detectors that achieve low end-to-end latency without sacrificing accuracy remains a key unresolved issue.

\textbf{Energy-efficient Computing Challenges.} Ensuring energy efficiency in neuromorphic object detection models is crucial, meeting mobile device constraints in real-world applications. Some studies~\cite{cordone2022object, su2023deep} reveal that SNNs are energy-efficient models that could achieve comparable performance to ANNs with the same architecture while consuming less energy. While many SNN-based models excel in energy efficiency for various tasks (e.g., classification~\cite{shrestha2018slayer}, optical estimation~\cite{paredes2019unsupervised}, and motion segmentation~\cite{parameshwara2021spikems}), they still largely operate on image-like representations rather than processing events asynchronously during training. The main reason is that current SNN training strategies are tightly coupled to timestep-based computation. They often convert event streams into frame-like tensors and rely on surrogate-gradient training, which requires discrete simulation steps. However, operating directly on raw event streams is where SNNs can effectively achieve low latency and low power on neuromorphic hardware. Therefore, designing asynchronous event-driven SNNs for object detection remains a key open challenge.

\begin{table*}
   \caption{Comparison of available object detection datasets using neuromorphic cameras.}
   \label{tab:object_detection_dataset}
   \scriptsize
   \vspace{-0.35cm}
   \begin{center}
   \renewcommand{\arraystretch}{1.10}
   \setlength{\tabcolsep}{1.05mm}{
   \begin{tabular}{lccccccccccccc}
   \toprule
        Dataset & Year  & Venue & Type & Scene & Resolution & Modality  & Label & Classes & Frames  & Boxes & Frequency  & High-speed & Low-light  \\
        \hline
        N-Caltech101~\cite{orchard2015converting} & 2015 & FNS & Real-world & Various & 304$\times$240 & Events & Manual & 100 & - & 8246 & 1 Hz & \ding{55} & \ding{55} \\
        PKU-DDD17-CAR~\cite{li2019event} & 2019 & ICME & Real-world & Driving  & 346$\times$260 & Events, Frames  & Manual  & 1 & 3155  & -  & 1 Hz  & \ding{55} & \ding{51}  \\
        PED Detection~\cite{miao2019neuromorphic} & 2019 & FNR & Real-world & Driving & 346$\times$260 & Events & Manual & 1 & 4670 & -  & 50 Hz & \ding{55} & \ding{55} \\
        Gen1 Detection~\cite{de2020large} & 2020  & arXiv & Real-world & Driving & 304$\times$240 & Events & Manual & 2 & 100.8k & 255k & 1, 4 Hz  & \ding{55} & \ding{51} \\
        1Mpx Detection~\cite{perot2020learning} &  2020 & NIPS & Real-world & Driving  & 1280$\times$720 & Events & Automated & 3 & 3.1M  & 25M & 60 Hz  & \ding{55} & \ding{51}  \\
        DAD~\cite{liu2021attention} & 2021  & ICIP & Real-world & Driving & 346$\times$260 & Events, Frames & Manual & 1 & 6456 & -  & 1 Hz  & \ding{55} & \ding{51}  \\
        PKU-Vidar-DVS~\cite{li2022retinomorphic} & 2022 & AAAI & Real-world & Various & Hybrid & Events, Spikes & Manual & 9 & 99.6K & 215.5K & 50 Hz & \ding{51} & \ding{51} \\
        Quadrotors Flying~\cite{iaboni2022event} & 2022 & Sensors & Real-world & Flying  & 640$\times$480 & Events & Manual & 1 & 10k & -  & - & \ding{55} & \ding{51}  \\
        N-AU-DR~\cite{andersen2022event} & 2022 & ECC & Hybrid & Flying & 240$\times$180 & Events, Frames & Manual & 1 & - & 465k & 120 Hz & \ding{51} & \ding{51} \\
        DSEC-Fusing~\cite{tomy2022fusing} & 2022 & ICRA & Real-world & Driving & 640$\times$480 & Events, Frames & Automated & 3 & - & 131.9k & - & \ding{55} & \ding{51} \\
        DSEC-MOD~\cite{zhou2023rgb} & 2023 & ICRA & Real-world & Driving & 640$\times$480 & Events, Frames & Automated & 8 & - & 13.3k & - & \ding{55} & \ding{51}  \\
        EventKITTI~\cite{liang2022global} & 2023 & MFI & Simulated & Driving & 1742$\times$375 & Events, Frames & Manual & 8 & \textgreater 7481 & - & - & \ding{55} & \ding{55} \\
        DSEC-Det~\cite{liu2023enhancing} & 2023 & arXiv & Real-world & Driving & 640$\times$480 & Events, Frames & Manual & 8 & 63.9k & 208k & - & \ding{55} & \ding{51}  \\
        Aqua-Eye~\cite{luo2023transcodnet} & 2023 & RAL & Real-world & Underwater & 346$\times$260 & Events, Frames & Manual & 5 & 6497 & 17.5k & - & \ding{55} & \ding{51} \\
        PEDRo~\cite{boretti2023pedro} & 2023 & CVPRW & Real-world & Robotics & 346$\times$260 & Events, Frames & Automated & 1 & 27k & 43.3k & 25 Hz & \ding{55} & \ding{51} \\
        PKU-DAVIS-SOD~\cite{li2023sodformer} & 2023  & TPAMI & Real-world & Driving  & 346$\times$260 & Events, Frames & Manual & 3 & 276k & 1097.3k  & 25 Hz  & \ding{51} & \ding{51}  \\
        TUMTraf~\cite{cress2024tumtraf} & 2024 & arXiv & Real-world & Roadside & 640$\times$480 & Events, Frames & Manual & 7 & 4.1k & 50.5k & - & \ding{55} & \ding{51} \\
        DSEC-Detection~\cite{gehrig2024low} & 2024 & Nature & Real-world & Driving & 640$\times$480 & Events, Frames & Automated & 8 & 70.4k & 390k & 20 Hz & \ding{55} & \ding{51} \\
        SEVD~\cite{aliminati2024sevd} & 2024 & CVPRW & Simulated & Flying & 1280$\times$960 & Events, Frames & Automated & 3 & - & 9M & 10 Hz & \ding{55} & \ding{51} \\
        eTraM~\cite{verma2024etram} & 2024 & arXiv & Real-world & Driving & 1280$\times$720 & Events & Annotated & 3 & - & 2M & 2 Hz & \ding{55} & \ding{51} \\
         NU-AIR~\cite{iaboni2023nu} & 2025 & IJCV & Real-world & Flying & 640$\times$48 & Events & Manual & 2 & - & 93.2k & 30 Hz & \ding{55} & \ding{55} \\
        EvDET200K~\cite{wang2025object} & 2025 & CVPR & Real-world & Various & 1280$\times$720 & Events & Manual & 10 & - & 202.3k & - & \ding{55} & \ding{51} \\
        EV-UAV~\cite{chen2025event} & 2025 & CVPR & Real-world & Tiny UAVs & 346$\times$260 & Events & Manual & 1  & - & 2.3M & -  & \ding{55} & \ding{51} \\
        \bottomrule
    \end{tabular}}
\end{center}
\vspace{-0.35cm}
\end{table*}

\section{Datasets}\label{sec:dataset}
This section first reviews existing neuromorphic object detection datasets from two perspectives. Then, we discuss the current status and the opportunities of these datasets.

\subsection{Simulated Datasets}\label{sec:simulated}
In the deep learning era, large-scale simulated event-based vision datasets~\cite{kang2021retinomorphic, aliminati2024sevd} are crucial for enhancing learning-based algorithms through data-driven strategies. 

\textbf{Video-to-event Datasets}. In general, neuromorphic camera simulators mimic the actual sampling process of real cameras. For example, Rebecq~\emph{et al.}~\cite{rebecq2018esim} first develop an event camera simulator (i.e., ESIM), which stands out for providing various features including event streams, RGB frames, optical flow, depth, and camera pose. Hu \emph{et al.}~\cite{hu2021v2e} design a realistic V2E to convert videos into event streams. ICNS~\cite{joubert2021event} is presented to generate high-fidelity events via simulating noise and estimating the latency. These simulators have enabled the creation of large-scale synthetic event-based datasets, effectively reducing annotation labor costs and advancing learning algorithms. A representative example is EventKITTI~\cite{liang2022global}, a dataset for automotive object detection generated by converting the KITTI video data~\cite{geiger2012we} using the ESIM tool. Other works, such as SEVD~\cite{aliminati2024sevd}, event-based VisDrone-VID~\cite{hannan2024event}, N-MobiFace~\cite{himmi2024ms}, and N-YoutubeFaces~\cite{himmi2024ms}, also follow this synthetic paradigm. Nevertheless, these simulated datasets~\cite{aliminati2024sevd, jing2024mvt, hannan2024event, himmi2024ms} may fail in capturing high-fidelity light changes in real-world challenging scenarios, which are precisely the situations where event cameras excel.

Current video-to-events synthetic strategies~\cite{iaboni2022event, liang2022global, li2022asynchronous} convert video datasets from the image domain to the event domain, but it may be hard to generate challenging datasets for converting any scenes. To address this limitation, physics-based event camera simulators (e.g., PECS~\cite{han2024physical}) can be integrated with unreal engine platforms such as AirSim\footnote[3]{\url{https://microsoft.github.io/AirSim/}}, CARLA\footnote[4]{\url{https://carla.readthedocs.io/en/latest/}}, and NVIDIA Isaac\footnote[5]{\url{https://developer.nvidia.com/isaac}}. These integrations enable the generation of event data within interactive 3D simulated environments, creating a pathway toward highly realistic datasets for challenging scenarios. Such simulated platforms are critical for advancing embodied neuromorphic agents by supporting the development and evaluation of closed-loop perception–action systems. In addition, some generative models (e.g., NeRF~\cite{tonderski2024neurad}, 3DGS~\cite{han2024event}, and diffusion model~\cite{croitoru2023diffusion}) could be employed to generate event data for various scenes, along with other modal data like frames and point clouds.

\textbf{Monitor Recording Datasets}. Some event-based simulated datasets (e.g., N-MNIST~\cite{orchard2015converting}, N-Caltech101~\cite{orchard2015converting}, CIFAR10-DVS~\cite{li2017cifar10}, and N-ImageNet~\cite{kim2021n}) use event cameras to capture images from popular frame-based datasets displayed on an LCD monitor. These monitor-recorded datasets help reduce costs and promote the development of event-based algorithms. Most are designed for object classification tasks~\cite{zhao2023spireco}, with N-Caltech101 being one of the few suitable for object detection. Take N-Caltech101 for example, each static image is recorded by a moving event camera positioned in front of an LCD monitor. Although these datasets are captured using real neuromorphic cameras, the motion speed and lighting conditions are constrained by the static nature of the original images, making it difficult to replicate the complexity of real-world scenarios. As a result, these simulated datasets limit the real-world application of neuromorphic object detection models in challenging conditions.

\subsection{Real-world Datasets}\label{sec:real_world}
There is a limited availability of object detection datasets using neuromorphic cameras, and they can be broadly divided into two categories from a modal perspective.

\textbf{Single-modal Datasets}. The first category~\cite{miao2019neuromorphic, de2020large, perot2020learning, iaboni2023nu, verma2024etram} refers to single-modal datasets typically constructed by using a DVS camera for building object detection datasets. For instance, the Gen1 Detection dataset~\cite{de2020large} and the 1Mpx Detection dataset~\cite{perot2020learning} offer extensive annotations for object detection within driving scenarios, utilizing a single DVS camera with varying spatial resolutions. The Quadrotor Flying dataset~\cite{iaboni2022event} is an open-source event-based object detection, featuring multiple quadrotors flying simultaneously at varying speeds and challenging light conditions. The NU-AIR dataset~\cite{iaboni2023nu} is the first event-based aerial drone dataset for pedestrian and vehicle detection in urban environments. While some specific scenes may perform well with single-modal DVS events, achieving high-precision object detection and capturing fine textures becomes challenging, particularly in stationary or extremely slow-motion scenarios. In short, event cameras excel at capturing dynamic motion but lack the ability to provide rich texture details like RGB images. Consequently, the number of labeled bounding boxes $C_{b}$ in event-based datasets is often limited to avoid producing an excessively small detection performance score.

\textbf{Multimodal Datasets}. The second category~\cite{li2019event, liu2021attention, li2022retinomorphic, zhou2023rgb, liu2023enhancing, luo2023transcodnet, li2023sodformer, cress2024tumtraf, magrini2024neuromorphic} involves \emph{multimodal datasets} that record multiple sensing streams to improve the accuracy of object detection. The PKU-DDD17-CAR dataset~\cite{li2019event}, the DAD dataset~\cite{liu2021attention}, and the Aqua-Eye dataset~\cite{luo2023transcodnet} offer 3,155, 4,670, and 6,497 hybrid pairs of events and frames, respectively. Yet, these small-scale multimodal datasets may be hard to provide effective data support for model training. Besides, the annotations for DSEC-Fusing~\cite{tomy2022fusing}, DSEC-MOD~\cite{zhou2023rgb}, DSEC-Det~\cite{liu2023enhancing}, and DSEC-Detection~\cite{gehrig2024low} are based on the widely used the DSEC dataset~\cite{gehrig2021dsec} for object detection. The PKU-Vidar-DVS dataset~\cite{li2022retinomorphic} is a large-scale multimodal neuromorphic object detection dataset using a hybrid camera system consisting of DAVIS346~\cite{moeys2017sensitive} and Vidar Gen1~\cite{huang20231000}. The PKU-DAVIS-SOD dataset~\cite{li2023sodformer} is a large-scale and temporally long-term multimodal object detection dataset, which is recorded in real-world driving scenarios with abundant diversity in object categories, object scales, motion speeds, and light changes.

Overall, the real-world event-based object detection dataset is still in its early stages and is gradually expanding (see Table~\ref{tab:object_detection_dataset}). Compared with image or video data, current real-world datasets are difficult to match in terms of labeled scale, object category, spatial resolution, etc. From Table~\ref{tab:object_detection_dataset}, most datasets now provide multimodal data, typically including events and frames. The spatial resolution and the number of labeled bounding boxes are generally increasing. Moreover, datasets are gradually moving toward providing continuous annotations rather than sparse 1 Hz labels, supporting streaming object detection. Nevertheless, most existing datasets still primarily support single-task 2D object detection. Actually, these three aspects can be considered when building datasets as follows: (i) Developing large-scale event-based datasets with high spatial resolution, multi-modalities, and multi-tasks, such as nuScenes~\cite{caesar2020nuscenes}; (ii) Constructing event-based datasets that involve challenging scenarios with high-speed motion and low-light settings, such as the low-light LLE-VOS dataset~\cite{li2024event} using optical filters to control light exposure; (iii) Building event-based datasets that go beyond 2D object detection tasks, encompassing 3D object detection tasks as well.

\section{Evaluation Metrics}\label{sec:metric}
This section reviews the metrics for evaluating neuromorphic object detection, focusing on detection accuracy, computational efficiency, and energy consumption.

\subsection{Detection Accuracy}\label{sec:detection_accuracy}
Evaluation metrics~\cite{perot2020learning, li2023sodformer} for neuromorphic object detection are directly inspired by the average precision (AP) criteria commonly employed in the traditional frame-based object detection domain. AP is typically evaluated for detection accuracy on a category-specific basis, calculated separately for each object category. To provide a comprehensive comparison across all categories, the mean average precision (mAP)~\cite{everingham2010pascal}, computed by averaging AP values across all object categories, serves as the ultimate performance metric. The AP~\cite{lin2014microsoft} is derived from precision and recall using the following formulas:
\begin{eqnarray}
    \text{AP} = \int_{0}^{1} max\{p(r^{'}|r^{'} \geq r)\} dr,
\end{eqnarray}
where $r$ denotes the recall that is the proportion of actual positive instances correctly identified by the object detection model, while $p(r)$ represents the precision as a function of recall along the precision-recall curve. The values of recall and precision are determined using a predefined intersection over union (IoU) threshold. The IoU is defined as the ratio of the overlapping area between the predicted bounding box and the ground-truth bounding box to the area of their union.

Thus, the mAP is calculated as the average of AP values across all object categories as follows:
\begin{eqnarray}
    \text{mAP} = \frac{\sum_{i=1}^{C_{b}} AP(i)}{C_{b}}.
\end{eqnarray}

At present, MS COCO\footnote[6]{\url{https://github.com/cocodataset/cocoapi}} is the most widely used benchmark for evaluating event-based object detection methods. Evaluation protocols with a fixed IoU threshold can be sensitive to minor localization errors and fail to reflect performance across varied object sizes. In contrast, the MS COCO protocol~\cite{perot2020learning} provides a more robust assessment by averaging performance across multiple IoU thresholds (i.e., $\text{mAP}$, $\text{mAP}_{0.5}$, and $\text{mAP}_{0.75}$) and AP across different scales (i.e., $\text{mAP}_{S}$, $\text{mAP}_{M}$, and $\text{mAP}_{L}$). The primary motivation for this multi-threshold strategy is to provide a more comprehensive assessment of localization robustness. However, a fundamental temporal mismatch exists between the neuromorphic sensing mechanism and the prevailing evaluation paradigm. While neuromorphic cameras offer continuous visual streams, ground truth labels are typically generated from synchronized RGB or reconstructed intensity frames at a fixed rate (e.g., 30 Hz). Thus, most existing evaluation pipelines are sampled rather than continuous, assessing detection accuracy only at the specific timestamps where labels are available. This frame-based evaluation protocol lacks the granularity to assess the model's performance on the entire event stream between two labeled frames, potentially masking the high-temporal-resolution and low-latency benefits inherent in neuromorphic sensing.

\subsection{Computational Efficiency}\label{sec:inference_speed}
When evaluating the overall computational efficiency of event-based object detectors~\cite{liu2023motion, mitrokhin2018event}, two commonly utilized metrics are inference time and model size. These two metrics emphasize the importance of rapid processing and minimal resource requirements for real-world applications.

\textbf{Inference Time}. Inference time~\cite{li2022asynchronous} is a widely used metric for evaluating a model's computational latency, defined as the duration from receiving an input event bin to producing the detection results. It highlights that fair comparisons of inference time should be tested on the same hardware platform. Furthermore, evaluations of SNN-based object detectors should ideally be conducted on dedicated neuromorphic computing platforms, as conventional GPUs are not well-suited for SNNs. It is also worth noting that the overhead of segmenting the continuous event stream into bins is negligible compared with the model's inference time. In other words, this binning preprocessing latency is seldom accounted for in existing neuromorphic object detectors. In the event-based vision community, the unit of measurement is typically milliseconds (ms) rather than frames per second (FPS). In general, lower inference times are desirable, particularly for low-latency neuromorphic cameras used in agile robotic applications.

\textbf{Model Size}. The size of the learning-based method~\cite{liu2023motion} is commonly quantified by the number of model parameters. The unit of model size is usually in megabytes (MB). Smaller model sizes are advantageous for computational efficiency, leading to faster inference time and suitability for deployment on resource-constrained devices. For non-deep learning techniques, assessing the algorithm's efficiency often involves evaluating the model's complexity.

\subsection{Energy Consumption}\label{sec:energy_consumption}

Power consumption is a crucial metric in the neuromorphic community, especially when working with SNN-based models. Current SNN-based object detectors are rarely deployed on real neuromorphic hardware for full energy measurement. In fact, prior works~\cite{kim2020spiking, chakraborty2021fully, park2024701} often estimate energy consumption theoretically using the number of accumulate (AC) and multiply-accumulate (MAC) operations, assuming a specific chip technology~\cite{horowitz20141}. An AC operation reflects simplified one-bit additions (e.g., spike accumulation in SNNs), while a MAC operation refers to multiply-add operations commonly used in ANNs. By multiplying these operational costs by the number of operations, this approach yields a theoretical computational energy estimate for SNNs. While these metrics are a staple of SNN-based object detectors used to demonstrate energy efficiency, they are equally relevant to ANN-based architectures. However, such theoretical analysis can reveal potential energy advantages of SNNs, it is not equivalent to actual system-level energy. In short, theoretical computational energy is only a part of the total system energy consumption. A complete analysis should account for various hardware factors. For instance, the communication energy (e.g., transmitting and routing events across multi-chip neuromorphic systems) represents a highly critical factor, often dominating the overall energy consumption over pure computational operations. Besides, neuromorphic chips consume mainly static power in the absence of spikes, whereas clock-driven designs benefit less from sparsity. Additionally, spiking neurons introduce overhead because their states (e.g., membrane potentials) must be maintained and updated over time. Updating these potentials at each time step incurs extra memory read and write energy. Ignoring these contributions can lead to misleading conclusions about SNN energy efficiency. Thus, experimental validation on actual neuromorphic platforms is critically needed to confirm these energy-efficiency benefits in real-world applications.

\begin{table}
   \caption{Summary of common metrics for event-based object detection.}
   \label{tab:object_detection_metric}
   \scriptsize
   \begin{center}
   \renewcommand{\arraystretch}{1.10}
   \setlength{\tabcolsep}{0.80mm}{
   \begin{tabular}{l c p{6.20cm}}
   \toprule
        Metric & Unit & \multicolumn{1}{c}{Description}  \\
        \hline
        $\text{mAP}$ & - &  mAP averaged over ten IoUs: \{0.5:0.05:0.95\}.  \\
        $\text{mAP}_{0.5}$ & - & mAP at a fixed IoU=0.50.\\
        $\text{mAP}_{0.75}$ & - & mAP at a fixed IoU=0.75.\\
        $\text{mAP}_{S}$ & - & mAP for small objects of area smaller than 32$^{2}$.\\
        $\text{mAP}_{M}$ & - & mAP for objects of area between 32$^{2}$ and 96$^{2}$.\\
        $\text{mAP}_{L}$ & - & mAP for large objects of area bigger than 96$^{2}$.\\
        \hline
        Inference time & ms & The duration from the input to the output per event bin.  \\
        Model size & MB & The number of parameters for the learning-based model.\\
        \hline
        \multirow{2}{*}{Power consum.} & \multirow{2}{*}{pJ} & The energy consumption of the SNN model through AC and MAC operations in a neuromorphic chip. \\
        \bottomrule
    \end{tabular}}
\end{center}
\vspace{-0.30cm}
\end{table}

\subsection{Analysis of Evaluation Metrics} \label{sec:analysis_metrics}

Current evaluation practices for neuromorphic object detection predominantly rely on metrics inherited from conventional frame-based computer vision. As summarized in Table~\ref{tab:object_detection_metric}, these established measures provide a foundation for comparative analysis but lack specificity for assessing neuromorphic object detection. In other words, while mAP is a widely used metric for evaluating detection accuracy and performs well in most conventional settings, it may not be optimal for assessing neuromorphic object detectors designed for low-latency applications. For example, detecting objects such as vehicles or pedestrians with minimal delay is critical in real-world autonomous driving applications. This is because mAP treats detections at different timestamps equally, without accounting for the temporal aspect of detection latency~\cite{mao2019delay}. Therefore, a more comprehensive metric that jointly considers both detection accuracy and temporal responsiveness would better reflect the performance of neuromorphic detectors under stringent low-latency requirements. Although such metrics have rarely been explored, developing new evaluation metrics is essential for improving both accuracy and reaction speed in low-latency event-based object detection.

Beyond standalone metrics such as accuracy, inference speed, and power consumption, the energy-delay product (EDP)~\cite{sharad2013spin} serves as a highly relevant figure-of-merit for evaluating event-driven systems. The EDP can be defined as the product of energy consumption and the processing latency, which quantitatively captures the critical efficiency trade-off between processing speed and power consumption. While the EDP is the gold standard for neuromorphic systems, most reported computational speeds usually measure the inference time of detection models (e.g., the forward-pass execution of the neural network) rather than the true system-level latency (i.e., the end-to-end latency from event data acquisition to final detection output). Similarly, power consumption is typically a theoretical algorithmic estimation (e.g., using AC and MAC operation counts) for a specific neuromorphic chip rather than measured directly on physical neuromorphic hardware. Future benchmarking efforts should also include the reporting of EDP to enable a more comprehensive and meaningful perspective on system-level performance evaluation.

\section{Neuromorphic Object Detection Methods}\label{sec:method}
This section will review existing neuromorphic object detection algorithms, covering event representation, temporal modeling, multimodal fusion, asynchronous processing, low-latency processing, and energy-efficient computing.

\subsection{Event Representation}\label{sec:representation}
Synchronous event-based object detectors~\cite{li2022asynchronous, zubic2023chaos, peng2023get, zubic2024state} often convert asynchronous events into 2D image-like representations before feeding them into frame-based object detectors. These event representations are designed to be compatible with frame-based methods~\cite{gehrig2019end} while maximizing object detection performance. Specifically, the continuous event stream is first divided into event bins with a constant window or a constant event number. Then, a kernel function is designed to map each event bin into an event embedding, which should ideally preserve the spatiotemporal information to optimize detection performance. The kernel function usually adopts a handcrafted operation or an end-to-end neural network. Existing event representations can be broadly classified into four categories according to the kernel function as follows.

\textbf{Event Images}. Each event bin is converted in a simple way by counting events or accumulating polarity pixel-wise into an image~\cite{maqueda2018event, gallego2020event, silva2024event}. For example, Jiang~\emph{et al.}~\cite{jiang2019mixed} directly project each event bin into a binary image before feeding it into the YOLOv3 network. Li~\emph{et al.}~\cite{li2023sodformer} map each event bin into a two-channel image to achieve an accuracy-speed trade-off, where the two channels separately encode the positive and negative polarity events and are then stacked to form the image. The widespread use~\cite{barua2016direct, chen2018pseudo} of event images is evident due to their straightforward method, which swiftly enables compatibility between events and frame-based technology. 
While pre-processing event images improves efficiency, it constrains the utilization of event timing information to its fullest extent. In essence, these operations have difficulty in effectively leveraging the spatiotemporal attributes of events.

\textbf{Handcrafted Features}. Spatiotemporal handcrafted features are often designed through the expertise-driven design of kernel functions to convert asynchronous events. Some studies~\cite{perot2020learning, ryan2021real, lenz2020event, tomy2022fusing, dadson2025marine} first utilize handcrafted features (e.g., voxel grids~\cite{zhu2019unsupervised} and time surfaces~\cite{lagorce2016hots}), then feed directly into the frame-based object detector. For example, Zhu~\emph{et al.}~\cite{zhu2019unsupervised} design the bilinear sampling kernel to convert events into voxel grids, and then Perot~\emph{et al.}~\cite{perot2020learning} use this voxel grid representation as the input of the designed ConvLSTM network for object detection. Besides, some works~\cite{mitrokhin2018event, chen2019multi, cordone2022object, zubic2023chaos, peng2023better, gehrig2023recurrent, wang2023dual, zubic2024state} focus on designing new handcrafted features (e.g., ERGO~\cite{zubic2023chaos} and hyper histogram~\cite{peng2023better}) specifically designed for object detection. For instance, Mitrokhin~\emph{et al.}~\cite{mitrokhin2018event} present a novel time-image representation that uses information about the temporal component of the event stream. Peng~\emph{et al.}~\cite{peng2023better} propose an efficient hyper histogram representation that preserves both the polarity and temporal information of the event stream. Zubi{\'c}~\emph{et al.}~\cite{zubic2023chaos} design a 12-channel event representation through gromov-Wasserstein optimization (ERGO-12) for object detection. 
Numerous experiments have shown that manually crafted features can achieve a good balance between inference speed and accuracy. However, their design process is time-consuming and heavily dependent on prior knowledge.  

\textbf{ANN-based Learned Representations}. Deep learning techniques have only recently been applied to learn end-to-end event representations in a data-driven manner. Some works aim to design general event representations (e.g., EST~\cite{gehrig2019end} and Matrix-LSTM~\cite{cannici2020differentiable}) for various event-based vision tasks, and these representations have also been validated for object detection. For example, Gehrig~\emph{et al.}~\cite{gehrig2019end} first utilize a multi-layer perceptron (MLP) to learn a dense representation end-to-end directly from asynchronous events. Cannici~\emph{et al.}~\cite{cannici2020differentiable} design a unique differentiable recurrent surface, termed Matrix-LSTM, employing a grid of long short-term memory cells to process raw events. Furthermore, there are ongoing attempts to learn end-to-end spatiotemporal representations (e.g., event embedding~\cite{li2022asynchronous} and EventPilars~\cite{wang2023dual}) from the event stream to maximize the performance of object detection tasks. For instance, Xu~\emph{et al.}~\cite{xu2021polar} employ an MLP with two hidden layers to learn an end-to-end event tensor. Li~\emph{et al.}~\cite{li2022asynchronous} present a temporal attention convolutional network that applies 1D causal convolution to process event sequences at each pixel, ultimately merging them into an attention embedding. Wang~\emph{et al.}~\cite{wang2023dual} introduce EventPillars, a learnable representation treating each event as a point, utilizing a 2D convolutional network to encode each pillar. Hamaguchi~\emph{et al.}~\cite{hamaguchi2023hierarchical} design an event sparse cross-attention module aimed at encoding event streams directly into dense memory cells with minimal information loss. Current ANN-based learned representations are the predominant choice for neuromorphic vision, consistently demonstrating state-of-the-art performance in various tasks.

\begin{table*}[htbp]
  \caption{Comparison of existing event representations on some representative neuromorphic object detectors.}
  \vspace{-0.30cm}
   \label{tab:event_representations}
   \scriptsize
	\begin{center}
        \renewcommand{\arraystretch}{1.15}
		\setlength{\tabcolsep}{0.96mm}{
			\begin{tabular}{l  l c l cc l}
				\toprule
				\multicolumn{1}{c}{Type} & \multicolumn{1}{c}{Event representation} & Dimensions & \multicolumn{1}{c}{Description} & Temporal & Polarity & \multicolumn{1}{c}{Typical object detectors} \\
				\hline
				\multirow{3}*{Event images} & Binary image~\cite{rebecq2017real}  & 2$\times H \times W$ & Two-channel image with width $W$ and height $H$ & \ding{55} & \ding{51} & ADF~\emph{et al.}~\cite{liu2021attention}, SODFormer~\cite{li2023sodformer} \\
				& Event count image~\cite{maqueda2018event}  & 2$\times H \times W$ & Rate-based image of event counts   & \ding{55} & \ding{51} & Chen~\emph{et al.}~\cite{chen2018pseudo}, FAGC~\cite{cao2021fusion}  \\
				& Grayscale image~\cite{wang2023spike} & $ H \times W$ & Binarized grayscale image after filtering & \ding{51}  &  \ding{51} & Wang~\emph{et al.}~\cite{wang2023spike} \\
                \hline
                \multirow{9}*{\shortstack{Handcrafted \\ features}} & Voxel grid~\cite{zhu2019unsupervised} & $ C \times H \times W$ & Voxel grid summing events with channel $C$ & \ding{51}  & \ding{55} & RED~\cite{perot2020learning}, FPN-fusion~\cite{tomy2022fusing}  \\
                & Voxel cube~\cite{cordone2022object} & $ C \times T \times H \times W$  & 4D voxel grid tensor with timestep $T$  & \ding{51} & \ding{55} & Spiking DenseNet~\cite{cordone2022object}, SFOD~\cite{fan2024sfod} \\
                & TE embedding~\cite{yu2024spikingvit} & $ T \times H \times W$ & Temporal extension embedding & \ding{51} & \ding{55} & SpikingViT~\cite{yu2024spikingvit} \\
                & Hyper histogram~\cite{peng2023better} &  4$ B \times H \times W$ & 3D temporal histograms with $B$ event bins & \ding{51} & \ding{51} & Peng~\emph{et al.}~\cite{peng2023better} \\
                & ERGO~\cite{zubic2023chaos} & 12$\times H \times W$ & Using Gromov-Wasserstein Discrepancy & \ding{51}  & \ding{51} & Zubi{\'c}~\emph{et al.}~\cite{zubic2023chaos}  \\
                & TORE volume~\cite{baldwin2022time} & 2$ K \times H \times W$ & 4D voxel grid of last $K$ timestamps & \ding{51}  & \ding{51} & Zubi{\'c}~\emph{et al.}~\cite{zubic2023chaos} \\
                & SAE~\cite{chen2019multi} & $ H \times W$ & Time surface of active events & \ding{51}  & \ding{51} & Chen~\emph{et al.}~\cite{chen2019multi}  \\
                & TAF~\cite{liu2023motion} & $ C \times H \times W$ & Event tensors with variable length & \ding{51}  & \ding{51} & AED~\cite{liu2023motion} \\
                & SSR~\cite{messikommer2020event} & $ C \times H \times W$ & Sparsely updating with each new event & \ding{51}  & \ding{55} & Messikommer~\emph{et al.}~\cite{messikommer2020event} \\
                \hline
                \multirow{6}*{ANN-based} & EST~\cite{gehrig2019end} & 2 $\times C \times H \times W$ & Learning an event spike tensor using MLPs & \ding{51}  & \ding{51} & Xu~\emph{et al.}~\cite{xu2021polar}, Zhou~\emph{et al.}~\cite{zhou2023end} \\
                & Matrix-LSTM~\cite{cannici2020differentiable} & $ C \times H \times W$ & Learning an event tensor using LSTMs &  \ding{51}  & \ding{51} & ASTMNet~\cite{li2022asynchronous}, EAS-SNN~\cite{wang2024eas} \\
                & Event embedding~\cite{li2022asynchronous} & $ C \times H \times W$ & Learning an event tensor using TACN & \ding{51}  & \ding{51} & ASTMNet~\cite{li2022asynchronous} \\
                & Group token embedding~\cite{peng2023get} &  $ G \cdot C \times  \frac{H}{P} \times \frac{W}{P} $ & $G$ Group tokens using $P$ $\times$$P$ patches and MLPs & \ding{51}  & \ding{51} & GET~\cite{peng2023get} \\
                & Reconstructed image~\cite{rebecq2019events} &  $ H \times W$ & Intensity image using E2VID algorithm  & \ding{51}  & \ding{55} & Hannan~\emph{et al.}~\cite{hannan2024event}  \\
                & SET~\cite{guo2024spatio} & 2$\times H \times W$ & Learning sparse event tensors using LSTMs & \ding{51} & \ding{51} & STAT~\cite{guo2024spatio} \\
                \hline
                \multirow{3}*{SNN-based} & Spike image~\cite{li2019event} & 2 $ \times H \times W$ & Two-channel image using the LIF model & \ding{51}  & \ding{51} & JDF~\cite{li2019event} \\
                & Leaky surface~\cite{cannici2019asynchronous} & $H \times W$ & Event image using the leak surface layer & \ding{51}  & \ding{55} & YOLE~\cite{cannici2019asynchronous}  \\
                & ARSNN~\cite{wang2024eas} & $ C \times H \times W$ & Event tensor using the recurrent SNNs & \ding{51} & \ding{51} & EAS-SNN~\cite{wang2024eas} \\
				\bottomrule
		\end{tabular}}
	\end{center}
    \vspace{-0.30cm}
\end{table*}

\textbf{SNN-based Learned Representations}. SNNs~\cite{zenke2021brain, neftci2019surrogate} are energy-efficient brain-inspired models encoding information in spatiotemporal dynamics and are theoretically capable of learning end-to-end representations. For instance, Li~\emph{et al.}~\cite{li2019event} first design a convolutional SNN to generate visual attention maps by firing rates of output neurons. Cannici~\emph{et al.}~\cite{cannici2019asynchronous} use the convolutional and max pooling layers to directly convert events into a unique leaky surface. Kugele~\emph{et al.}~\cite{kugele2021hybrid} utilize the leaky integrate-and-fire (LIF) model to generate an intermediate representation. Wang~\emph{et al.}~\cite{wang2024eas} present an adaptive sampling module with recurrent convolutional SNNs to learn a sparse spike representation. Due to their advantages in handling temporal dynamics and enhancing energy efficiency, SNNs are a compelling choice for learning end-to-end representations of spatiotemporal events.

As shown in Table~\ref{tab:event_representations}, we present an overview of the event representations for event-based object detection. While image-like representations are well-aligned with mainstream frame-based algorithms, they compromise the inherent sparsity and asynchronous nature of raw events. When designing event representations, the following three aspects should be considered: (i) Developing asynchronous representations that are updated incrementally with each event~\cite{messikommer2020event}; (ii) Designing pre-trained event representation models~\cite{sabater2022event} driven by large-scale datasets that are adaptable to multiple vision tasks and compatible with various modalities~\cite{li2025asynchronous} (e.g., frames and events); (iii) Changing sensor hardware and optical lenses~\cite{he2024microsaccade} to directly output high-contrast event representations in real-time, rather than relying on algorithmic post-processing.

\subsection{Temporal Modeling}\label{sec:temporal_modeling}
The event stream offers high-resolution and continuous-time attributes in the temporal domain~\cite{li2021asynchronous}. By leveraging the rich temporal cues, the detector can identify the same objects over time that closely resemble the current object and align them together~\cite{li2022asynchronous}. As a result, object detection algorithms using temporal cues can achieve higher accuracy and faster response times~\cite{gehrig2024low}. Existing temporal modeling strategies mainly include the following four aspects.

\textbf{Temporal Aggregation Operations.} Temporal aggregations~\cite{zhou2023rgb, li2022retinomorphic, hamaguchi2023hierarchical, wang2023dual, han2024real} often use attention mechanisms or concatenation operations to process multiple event temporal bins. For example, Zhou~\emph{et al.}~\cite{zhou2023rgb} present a temporal multi-scale aggregation module to leverage event bins. Li~\emph{et al.}~\cite{li2022retinomorphic} design a temporal aggregation operation using the attention mechanism for event bins. While these temporal aggregation techniques effectively balance accuracy and speed through simplified temporal modeling, they encounter difficulties in fully leveraging the rich temporal information within event streams, which somewhat constrains the model’s high-frequency inference capabilities.

\textbf{Recurrent-Convolutional Architectures.} Recurrent neural networks (RNNs)~\cite{yu2019review} and their variants are widely used for modeling long-term dependencies in event-based vision, leading to the development of various recurrent-convolutional architectures~\cite{perot2020learning, li2022asynchronous, wang2023dual, andersen2022event, silva2024recurrent, zhu2024spatio, jing2025esvt} for object detection. For instance, Perot~\emph{et al.}~\cite{perot2020learning} first introduce the ConvLSTM layers to leverage temporal cues for robust and high-accuracy object detection. Li~\emph{et al.}~\cite{li2022asynchronous} design a lightweight yet efficient inter-weaved recurrent-convolutional layer in SSD for temporal modeling. While recurrent-convolutional models effectively integrate information across event bins, their incorporation into feed-forward detection frameworks increases computational overhead. Consequently, deploying these recurrent-convolutional models on edge devices often requires carefully balancing accuracy and speed to ensure low-latency inference.

\begin{figure*}[t]
\centering
\includegraphics[width=\linewidth]{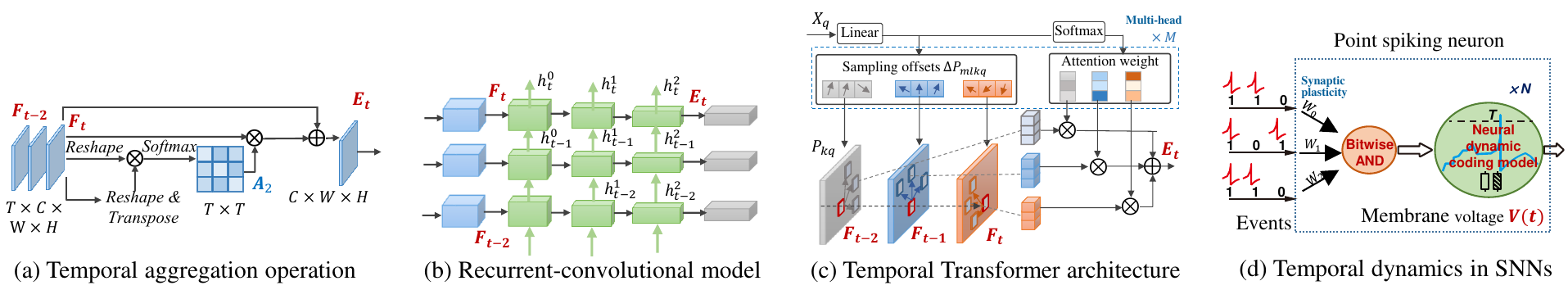}
\caption{Four typical temporal modeling architectures for processing event streams in object detection tasks. (a) Temporal aggregation operations using the attention mechanism~\cite{li2022retinomorphic}. (b) Recurrent-convolutional model (e.g., ConvLSTM~\cite{perot2020learning}) extracts spatiotemporal patterns. (c) Temporal Transformer~\cite{li2023sodformer} aggregates multiple features in the temporal domain. (d) Temporal information dynamics in SNNs~\cite{su2023deep}.}
\label{fig:temporal_modeling}
\vspace{-0.30cm}
\end{figure*}

\textbf{Temporal Transformers.} Global self-attention and parallel computation make the transformer highly effective for temporal modeling in sequential tasks~\cite{liu2024event, wang2022learning}. More recently, some approaches~\cite{gehrig2023recurrent, li2023sodformer} have utilized temporal transformers for temporal modeling in event-based object detection. Gehrig~\emph{et al.}~\cite{gehrig2023recurrent} integrate long short-term memory (LSTM) into recurrent vision transformers to exploit temporal cues for enhancing detection performance. Besides, Li~\emph{et al.}~\cite{li2023sodformer} present a temporal transformer module to use temporal cues from continuous event streams and adjacent frames, showing that this transformer enhances memory capabilities through effective long-term sequence modeling. Overall, transformers have emerged as a leading architecture in object detection due to their strong ability to capture temporal dependencies. However, neuromorphic object detectors using transformers generally incur higher computational costs compared with CNNs or lightweight RNNs, making optimization techniques such as model compression and acceleration essential for achieving low-latency inference in real-world applications.

\textbf{Temporal Dynamics in SNNs.} SNNs incorporate mechanisms such as membrane potentials with inherent leakage and refractory periods to accurately model the temporal dynamics of biological neurons~\cite{roy2019towards, he2024network, zhang2023gpu, zheng2024temporal, wu2022brain}. The leakage dynamics of the membrane potential enable neurons to integrate incoming weighted spikes over a finite time window, allowing them to detect temporal patterns, while the refractory period modulates firing behavior. These mechanisms allow SNNs to use temporal cues for accurate object detection~\cite{cordone2022object, su2023deep, wang2024eas}. For example, Cordone~\emph{et al.}~\cite{cordone2022object} directly train four SNN models to process event streams for object detection. Wang~\emph{et al.}~\cite{wang2024eas} present recurrent convolutional SNNs enhanced with temporal memory for event-based object detection. While SNNs haven't outperformed RNNs and temporal Transformers, their biological interpretability and low power consumption still hold promise for modeling temporal dependencies.

As illustrated in Fig.~\ref{fig:temporal_modeling}, we present four typical temporal modeling architectures in existing event-based object detectors~\cite{li2022retinomorphic, perot2020learning, li2023sodformer, su2023deep}. Various temporal architectures aim to enhance performance while maintaining low complexity. However, these models should consider deployment and acceleration on hardware. Besides, state space models~\cite{zubic2024state, liang2025esod} could be utilized to capture longer temporal dependencies.

\subsection{Multimodal Fusion}\label{sec:multimodal}
Leveraging the complementary benefits of event cameras with other modalities (e.g., RGB frames~\cite{li2023sodformer, gehrig2024low, lu2024flexevent, fan2025efficient}, infrared images~\cite{hu2020learning}, and LiDAR point clouds~\cite{li2021enhancing, wu2024moving}) is a popular strategy for high-accuracy and robust object detection in challenging scenarios. Most multimodal methods fuse events and RGB frames, which can be broadly classified into the following three categories.

\textbf{Data-level Fusion.} Data-level fusion~\cite{alremeithi2023event, luo2023transcodnet} occurs at the initial stage, combining raw events and degraded frames to produce higher-quality images or representations before feeding them into the object detector. For example, Alremeithi~\emph{et al.}~\cite{alremeithi2023event} implement a reconstruction step to generate enhanced images by fusing events and RGB images, making them more suitable for object detection tasks. Luo~\emph{et al.}~\cite{luo2023transcodnet} fuse events with RGB frames using a weighted average at the pixel level to enhance edge features, then input them into a CNN backbone to detect underwater transparently camouflaged organisms. While data-level fusion offers a quick and straightforward merging process, it requires managing disparities in data formats and resolutions.

\textbf{Feature-level Fusion.} Feature-level fusion for event-based object detection tasks typically involves extracting feature maps from two modalities and then fusing them using attention mechanisms~\cite{liu2021attention, cao2023chasing, liu2023enhancing, ma2025dsf} or concatenation operations~\cite{cao2021fusion, tomy2022fusing, cao2024embracing}. For instance, Liu~\emph{et al.}~\cite{liu2021attention} design a joint object detection network with an attention fusion module that utilizes features from both events and frames. Tomy~\emph{et al.}~\cite{tomy2022fusing} concatenate pyramidal event and frame features at the same scale before feeding them into RetinaNet. Feature-level fusion integrates complementary features within the intermediate layers of end-to-end neural networks. This strategy~\cite{li2023sodformer} achieves better performance than other methods (e.g., data-level and decision-level fusion), making it a mainstream choice for multimodal fusion in neuromorphic object detection.

\textbf{Decision-level Fusion.} Decision-level fusion~\cite{li2019event, chen2018pseudo, jiang2019mixed, jiang2023nighttime, cress2024tumtraf} combines the final outputs of independent object detectors from two modalities. For example, Jiang~\emph{et al.}~\cite{jiang2019mixed} present a confidence map fusion strategy combining the predicted results from two branches. Li~\emph{et al.}~\cite{li2019event} introduce the Dempster-Shafer theory to combine the outputs from events and frames from a CNN in a joint decision model. While post-processing fusion is useful for combining different modalities, its performance does not match that of data-level fusion or end-to-end feature-level fusion in most cases.

Most fusion strategies usually split event streams to match frame sampling and integrate synchronized streams, but the inference frequency is usually limited by frame sampling rates. Thus, exploring asynchronous fusion~\cite{li2023sodformer, gehrig2024low} is crucial for leveraging the high temporal resolution of event streams and meeting the low-latency detection requirements in high-speed scenarios (see Fig.~\ref{fig:asynchronous_processing}). While events and frames are the primary modalities, exploring additional modalities like infrared images or LiDAR point clouds in conjunction with events for multimodal object detection offers exciting possibilities~\cite{ta2023l2e}. Furthermore, investigating collaborative object detection with multiple agents using event cameras is also promising for a broader field of view and robust object detection.

\subsection{Asynchronous Processing}\label{sec:asynchronous}
Asynchronous processing~\cite{li2021graph, sabatier2017asynchronous} allows for individual event handling with low-latency. This asynchronous paradigm is essential for real-time inference in high-speed scenarios. 

\textbf{Sparse Convolutions}. Recently, sparse convolutions~\cite{messikommer2020event, durvasula2023ev, jack2020sparse} bring the spatiotemporal sparsity of events into high-performance CNNs, significantly reducing their computational complexity. For instance, Messikommer~\emph{et al.}~\cite{messikommer2020event} present a novel asynchronous neural network that processes events asynchronously for object detection. Jack~\emph{et al.}~\cite{jack2020sparse} develop a convolution operator for sparse events that utilizes only matrix products and addition during training. However, these sparse convolutions process events asynchronously with low latency but slightly reduced accuracy.

\textbf{Graph Neural Networks}. Some event-based vision models~\cite{schaefer2022aegnn, gehrig2022pushing, gehrig2024low, jeziorek2023memory, sun2023event, jeziorek2024optimising, wu2024egsst, verma2026event} have adopted graph neural networks (GNNs) to process events as inherently sparse graphs for object detection. For instance, AEGNN~\cite{schaefer2022aegnn} is the first attempt at a novel event-by-event processing paradigm that utilizes standard GNNs to handle events asynchronously. Despite early work showing impressive computational reductions, their accuracy is still limited by the small scale and shallow depth of frameworks. Gehrig~\emph{et al.}~\cite{gehrig2022pushing} overcome this shortage with EAGR, a novel GNN-based architecture that enables scaling the depth and complexity of models while maintaining low computational costs. Subsequently, they further present GAGr~\cite{gehrig2024low}, a hybrid object detector that combines a standard CNN for frames and an efficient asynchronous GNN for event data. Li~\emph{et al.}~\cite{li2025asynchronous} propose a novel asynchronous collaborative graph representation (ACGR), which explores a unified graph framework for asynchronously processing frames and events with high performance and low latency.

\textbf{Event-driven SNNs}. Combining event-driven SNNs~\cite{shrestha2018slayer, zeng2025leveraging} with event cameras offers a promising path toward high-accuracy, low-latency, and energy-efficient neuromorphic systems. For example, the asynchronous spiking graph convolutional network (ASGCN)~\cite{zeng2025leveraging} processes raw neuromorphic data on an event-by-event basis, achieving competitive accuracy with extreme energy efficiency compared to frame-based methods. Currently, asynchronous event-driven SNNs are primarily explored for classification tasks and have only rarely been applied to complex regression tasks such as object detection. While they offer efficient asynchronous processing, their deployment on neuromorphic hardware still faces challenges in implementation and acceleration optimization.

Overall, event-driven SNNs remain in the early stages of development. They currently lag behind synchronous image-like models in both detection accuracy and multi-task capability. Future research on event-driven SNN-based object detectors should address three key aspects: i) Asynchronously processing individual events in unimodal tasks to achieve accuracy comparable to image-like representations; ii) Achieving asynchronous fusion of events and images to enable high-frequency inference~\cite{li2023sodformer} and leverage complementary strengths of each modality; iii) Optimizing computational speed and memory efficiency to enable low-latency object detection on resource-constrained neuromorphic devices~\cite{ziegler2025detection}.

\begin{table*}[t]
   \caption{Comparison of three representative energy-efficient spiking neural models for neuromorphic object detection.}
   \label{tab:energy_efficient_snns}
   \scriptsize
    \begin{center}
        \renewcommand{\arraystretch}{1.10}
        \setlength{\tabcolsep}{0.95mm}{
            \begin{tabular}{l cccc}
                \toprule
                \multirow{2}*{Method} & \multirow{2}*{ANN-to-SNN conversation~\cite{chakraborty2021fully, wang2023spike}} & \multirow{2}*{Directly-trained SNNs~\cite{su2023deep}} & \multicolumn{2}{c}{Hybrid ANN-SNN~\cite{li2019event, kugele2021hybrid, ahmed2024hybrid}} \\ 
                \cline{4-5} & & & One branch & Two branches \\
                \hline
                Modality & Unimodal & Unimodal & Unimodal & Multimodal \\
                Typical inputs & Events or spikes & Events or spikes & Events or spikes & Events + frames \\
                Training strategy & Train ANNs first, then convert to SNNs & Surrogate gradient & Surrogate gradient and backpropagation & Joint training for ANNs and SNNs \\
                Training complexity & Moderate, train and convert ANNs & Simple, directly train &  Moderate, train both ANNs and SNNs & High, train separately and later merge \\
                Detection accuracy & Low & Medium & High & Higher \\
                Time complexity & $O(T_{\text{c-SNN}}) \gg O(T_{\text{ANN}})$  & $O(T_{\text{c-SNN}}) > O(T_{\text{d-SNN}})$ & $O(T_{\text{ANN}}) + O(T_{\text{SNN}})$ & $\max(O(T_{\text{ANN}}), O(T_{\text{SNN}})))$ \\
                Energy-efficacy & Medium &  Low & Large, a trade-off & Larger, mainly ANNs \\
                Hardware deployment & Simple & Simple & Moderate, both ANNs and SNNs & High, hybrid dual processing\\
                Typical object detectors &  Spiking-YOLOv4~\cite{wang2023spike} & EMS-YOLO~\cite{su2023deep} & JDF~\cite{li2019event}, Soikat~\emph{et al.}~\cite{ahmed2024hybrid} & HMN~\cite{zhao2022framework} \\
                \bottomrule
        \end{tabular}}
    \end{center}
    \noindent $O(T_{\text{c-SNN}})$, $O(T_{\text{d-SNN}})$, $O(T_{\text{SNN}})$, and $O(T_{\text{ANN}})$ denote the time complexities for ANN-to-SNN conversion, directly-trained SNNs, pure SNN models, and ANN models, respectively.
    \vspace{-0.40cm}
\end{table*}

\subsection{Low-latency Processing}\label{sec:low_latency}
Efficient object detection models~\cite{gehrig2024low, falanga2019fast, shi2023identifying} aim to balance accuracy and computational speed, making them perfect for real-time applications. Algorithm optimization and hardware acceleration are two primary approaches for advancing low-latency event-based object detection.

\textbf{Algorithm Optimization.} Current optimization strategies for low-latency object detection involve lightweight models~\cite{fan2024dense, el2023high, liu2023motion, yuan2023trainable, liang2021event, shi2023identifying} and event-by-event asynchronous processing networks. For example, Liu~\emph{et al.}~\cite{liu2023motion} present a lightweight yet effective model to achieve motion-robust and high-speed object detection. Gehrig~\emph{et al.}~\cite{gehrig2024low} combine frames using CNNs and events using GNNs, showing the potential of event cameras for accurate and low-latency object detection in edge-case scenarios. Ziegler~\emph{et al.}~\cite{ziegler2025event} present a low-latency ball detection system for a table tennis robot, designing an event representation that allows event-by-event updates. The real bottleneck may not be in processing image-like representations, but rather in the lack of parallel hardware capable of handling event-by-event data with low-latency. Conventional computing architectures (e.g., GPUs) are ill-suited for this asynchronous paradigm, as they are optimized for batch processing. Overcoming this limitation will require the development of novel neuromorphic processors that natively support event-by-event computation.

\textbf{Hardware Acceleration.} The hardware platforms~\cite{qiu2023highly, ziegler2025detection, wang2022ev, crafton2021hardware, jonathan2024embedded, ren2025neuromorphic, przewlocka2024poweryolo, li2024energy, sanaullah2024spike, silva2024end, chen2025low, lu2024real} used for deploying object detection algorithms include GPUs, FPGAs, and neuromorphic computing chips. For example, Wang~\emph{et al.}~\cite{wang2022ev} develop a catching system for fast-flying ping-pong balls on an embedded GPU platform, achieving an 81\% success rate across different target locations and velocities up to 13 m/s. Qiu~\emph{et al.}~\cite{qiu2023highly} present a pseudo-quantization scheme for SNNs, achieving a 118$\times$ speedup on GPU for object detection and enabling 800 FPS on FPGA platforms. Li~\emph{et al.}~\cite{li2021recent} develop a high-speed bullet detection system using a spiking camera on an FPGA platform, capable of real-time detection of a 150 km/h bullet within a 4.2-meter range. In fact, hardware acceleration efforts are fewer compared to algorithm optimization, yet combining event cameras with hardware acceleration is crucial for achieving low-latency processing.

Overall, current neuromorphic object detection models mainly focus on lightweight architectures to achieve real-time processing. However, there has been limited exploration of network compression techniques such as quantization~\cite{deng2020model}, pruning, knowledge distillation~\cite{li2023object}, and neural architecture search~\cite{silva2025chimera}, particularly for deployment on dedicated neuromorphic hardware accelerators. Consequently, it remains a major open question how to deploy detection models fast enough for neuromorphic computing chips such as SpiNNaker~\cite{furber2012overview}, TrueNorth~\cite{merolla2014million}, DYNAPs~\cite{moradi2017scalable}, Braindrop~\cite{neckar2018braindrop}, Tianjic~\cite{pei2019towards}, Loihi~\cite{davies2021advancing}, BrainScales~\cite{pehle2022brainscales}, and NorthPole~\cite{modha2023neural}, etc. A principal challenge here is that open-source tools for training, deployment, and inference tailored to these neuromorphic chips are still significantly scarcer compared to the well-established software ecosystem for conventional ANNs. Thus, developing efficient co-design acceleration methodologies that bridge the algorithmic and neuromorphic hardware domains remains a crucial open research direction.

\subsection{Energy-efficient Computing}\label{sec:energy_efficient}
Energy-efficient object detection methods~\cite{li2019event, su2023deep, wang2023spike, fan2024sfod, caccavella2023low, zhang2024automotive, hasssan2024spiking, iaboni2024event} usually integrate low-latency event cameras and low-power consumption SNNs, which can be broadly categorized into three categories (see Table~\ref{tab:energy_efficient_snns}).

\textbf{ANN-to-SNN Conversion.} The conversion from ANN to SNN~\cite{hu2023fast} aims to leverage the strengths of spiking neurons, such as lower power consumption and biological interpretability, while retaining the performance of traditional ANN models. Early ANN-to-SNN works~\cite{perez2013mapping, lee2016training} attempt to transform the continuous activation values in each layer into a series of spikes over time, enabling SNNs to capture temporal dynamics for classification tasks. Subsequent works~\cite{chakraborty2021fully, kim2020spiking, wang2023spike} design ANN-to-SNN conversion models for energy-efficient object detection, which typically require extended simulation time steps to better match the performance of the original ANN model. For instance, Kim~\emph{et al.}~\cite{kim2020spiking} propose a spike-based object detection model called Spiking-YOLO, which utilizes channel-wise normalization and signed neurons with imbalanced thresholds to achieve fast and accurate information transmission in deep SNNs. Wang~\emph{et al.}~\cite{wang2023spike} present Spiking-YOLOv4, a converted CNN model for fast and accurate object detection from event streams. Although those ANN-to-SNN conversion methods can quickly transform pre-trained models by converting activation functions into spiking neurons, their performance is inferior to directly-trained SNNs and requires longer simulation time steps.

\textbf{Directly-trained SNNs.} The direct training of SNNs has been significantly advanced through surrogate gradient strategies~\cite{neftci2019surrogate, zenke2021brain}, which provide differentiable approximations of spiking neuron dynamics during backpropagation. The pioneering work~\cite{wu2018spatio} demonstrates the effectiveness of surrogate gradients for training high-performance SNNs for classification tasks. Recent research~\cite{cordone2022object, barchid2023spiking, su2023deep, caccavella2023low, yuan2023trainable, bodden2024spiking, ziegler2025detection, wang2024eas, fan2024sfod, yu2024spikingvit, li2025brain, li2025yolo} has extended directly-trained SNNs to energy-efficient object detection. These methods typically require fewer time steps than conversion-based approaches while achieving competitive accuracy with their original ANN models. For example, Su~\emph{et al.}~\cite{su2023deep} propose EMS-YOLO, a directly-trained SNN-based model for object detection that outperforms ANN-SNN conversion methods, requiring only small time steps for real-time inference. Caccavella~\emph{et al.}~\cite{caccavella2023low} design a training scheme for SNNs deployed on the neuromorphic chip Speck to achieve low-power face detection. Indeed, by enabling simultaneous learning of temporal dynamics and spatial features during backpropagation, surrogate gradient methods have made SNNs increasingly practical and scalable. Yet, most current SNN-based object detectors rely on image-like representations as input to the SNN models, rather than implementing truly asynchronous event-driven processing. Moreover, their energy efficiency has seldom been assessed through direct measurements on neuromorphic hardware.

\begin{table*}[t]
   \caption{Comparison of state-of-the-art event-based object detectors on the Gen1 Detection dataset~\cite{de2020large}.}
   \vspace{-0.25cm}
   \label{tab:gen1_results}
   \scriptsize
   \begin{center}
   \renewcommand{\arraystretch}{1.10}
   \setlength{\tabcolsep}{1.30mm}{
   \begin{tabular}{l cc cccc ccccc}
        \toprule
        Method  & Type & Asynchronous & Representation & Backbone & Head  & Temporal & \# Param. (M) & mAP$_{50}$ & mAP$_{50:95}$    \\
        \hline
        RED~\cite{perot2020learning} & CNN-based & No & Voxel grid & CNN+RNN & SSD & ConvLSTM &  24.1  & - & 0.400 \\
        Crafton~\emph{et al.}~\cite{crafton2021hardware} & CNN-based & No & Binary image & CNN+RNN & YOLOv3 & ConvLSTM & - & - & 0.396 \\
        ASTMNet~\cite{li2021asynchronous} & CNN-based & No & Event embedding & CNN+RNN & SSD & Rec-Conv & 39.6 & - & 0.467 \\
        HMNet~\cite{hamaguchi2023hierarchical} & CNN-based & No & Dense memory cells & CNN & YOLOX & Latent Memory & - & - & 0.427 \\
        AED~\cite{liu2023motion} & CNN-based & No & TAF & CNN & YOLOX & No & 14.8 & - & 0.454 \\
        Peng~\emph{et al.}~\cite{peng2023better} & CNN-based & No & Hyper histogram & CNN & YOLOv5 & No & - & - & 0.470 \\
        CED~\cite{zhu2024spatio} & CNN-based & No & DTR & CNN & CDE & LRC & 19.8 & - & 0.472 \\
        E-detector~\cite{zhang2023detector} & Transformer-based & No & Reconstructed image & Transformer & DETR & No  & - & 0.483 & -\\
        GET~\cite{peng2023get} & Transformer-based & No & Group token & Transformer+RNN & YOLOX & ConvLSTM & 21.9 & - & 0.479 \\
        RVT-B~\cite{gehrig2023recurrent} & Transformer-based & No & 4D tensor & Transformer+RNN & YOLOX & LSTM & 18.5 & - & 0.475 \\
        LEOD-RVT-S~\cite{wu2024leod} & Transformer-based & No & 4D tensor & Transformer & YOLOX & Time-flip TTA & 9.9 & - & 0.487 \\
        S5-ViT-B~\cite{zubic2024state} & Transformer-based & No & 4D Tensor & Transformer+SSM & YOLOX & SSM & 17.5 & - & 0.474 \\
        STAT~\cite{guo2024spatio} & Transformer-based & No & SET & Transformer & YOLOX & TAM & - & 0.788 & 0.499 \\
        Zubi{\'c}~\emph{et al.}~\cite{zubic2023chaos} & Transformer-based & No & ERGO-12 & Transformer & YOLOv6 & No & - & - & 0.504 \\
        EvRT-DETR-B~\cite{torbunov2025evrt} & Transformer-based & No & 4D tensor & Transformer+RNN & DETR &  RNN & 57.1 & - & \textbf{0.527} \\
        SMamba~\cite{yang2025smamba} & SSM-based & No &  Voxel grid & SSM+RNN & YOLOX &  STCA & 16.1 & - & 0.504\\
        \hline
        Asynet~\cite{messikommer2020event} & CNN-based & Yes & Event histogram & Sparse CNN & YOLO & No & - & - & 0.129 \\
        AEGNN~\cite{sabater2022event} & GNN-based & Yes & Graph & GNN & YOLO & No & 20.0 & - & 0.163 \\
        EAGR-L~\cite{gehrig2022pushing} & GNN-based & Yes & Graph & GNN & YOLOX & No & - & - & 0.321 \\
        EGSST-E~\cite{wu2024egsst} & GNN-based & Yes & Graph & GNN+LinearViT & YOLOX & Yes & 12.3 & - & \textbf{0.496} \\
        \hline
        Ahmed~\cite{ahmed2024hybrid} & Hybrid SNN-ANN & No & 4D tensor & SNN & YOLOX & Yes & 6.6 & - & 0.340 \\
        SNN-MobileNet~\cite{cordone2022object} & SNN-based & No & Voxel cube & SNN & SSD & Yes & 24.3 & - & 0.147 \\
        SNN-DenseNet~\cite{cordone2022object} & SNN-based & No & Voxel cube & SNN & SSD & Yes & 8.2 & - & 0.189 \\
        EMS-YOLO~\cite{su2023deep} & SNN-based & No & Event histogram & SNN & YOLO & Yes & 6.2 & 0.547 & 0.267 \\
        FP-DAGNet~\cite{zhang2024automotive} & SNN-based & No & SBT & SNN & Multi-head & Yes & 22.0 & - & 0.223 \\
        Tr-Spiking-YOLO~\cite{yuan2023trainable} & SNN-based & No & Voxel grid & SNN & YOLO & Yes & - & 0.453 & -\\
        LT-SNN~\cite{hasssan2024spiking} & SNN-based & No & Time-sampled data & SNN & YOLOv2 & Yes & - & - & 0.298 \\
        SNN-CenterNet~\cite{bodden2024spiking} & SNN-based & No & Binary image & SNN & CenterNet & Yes & 13.0 & - & 0.229 \\
        SFOD~\cite{fan2024sfod} & SNN-based & No & Voxel cube & SNN & SSD & Yes & 11.9 & - & 0.321 \\
        SpikingViT-B~\cite{yu2024spikingvit} & SNN-based & No & TEE & SNN & YOLOX & Yes & 21.5 & 0.616 & 0.394 \\    
        EAS-SNN~\cite{wang2024eas} & SNN-based & No & ARSNN & SNN & YOLOX & Yes & 25.3 & 0.699 & 0.375 \\
        Wang~\emph{et al.}~\cite{wang2025adaptive} & SNN-based & No & Scaled timesurface & SNN & SSD & Yes & - & 0.578 & 0.304 \\
        CREST~\cite{mao2024crest} & SNN-based & No & MESTOR & SNN & YOLOv4 & Yes & 7.61 & 0.632 & 0.360 \\
        SpikeYOLO~\cite{luo2024integer} & SNN-based & No & Event histogram & SNN & YOLOv8 & Yes & 23.1 & 0.672 & 0.404 \\
        Li~\emph{et al.}~\cite{li2026spike} (M) & SNN-based & No & SDA-MTF & SNN & YOLOX & Yes & 25.3 & 0.731 & \textbf{0.434} \\
        \bottomrule
    \end{tabular}}
\end{center}
\noindent These bounded categories refer to image-like ANN-based models (i.e., CNNs, Transformers, and SSMs), asynchronous GNN-based models, and energy-efficient SNN-based models.
\vspace{-0.30cm}
\end{table*}

\textbf{Hybrid ANN-SNN Architectures.} In general, hybrid neural networks~\cite{zhao2022framework, aydin2024hybrid, li2026rethinking, shariff2025face, zhang2026event} combine the strengths of high-accuracy ANNs and energy-efficient SNNs in neuromorphic vision tasks, such as object detection~\cite{li2019event, kugele2021hybrid, ahmed2024hybrid}. ANNs excel at processing complex spatial information, while SNNs efficiently capture temporal dynamics using less energy. This complementary approach allows hybrid ANN-SNN architectures to achieve highly accurate object detection while maintaining energy efficiency, making them ideal for real-time applications in agile robotics and edge devices. Moreover, with advancements in hybrid vision sensors~\cite{brandli2014240, moeys2017sensitive, guo2023three, yang2024vision} and neuromorphic computing chips~\cite{pei2019towards}, there is an increasing trend towards designing these hybrid neural networks. One class~\cite{li2019event, kugele2021hybrid, ahmed2024hybrid} of methods uses SNNs for efficient encoding and feature extraction of spatiotemporal events, followed by different ANN backbones for various visual tasks. Another class~\cite{zhao2022framework} employs two branches with ANNs to extract features from frames and SNNs to directly process events, handling multimodal frame and event streams. However, training complexity and accuracy-speed trade-offs remain key considerations for designing hybrid ANN-SNN models on neuromorphic hardware.

Energy-efficient computing is ideal for edge resource-constrained devices such as micro-drones~\cite{andersen2022event} and space satellites~\cite{afshar2020event}. Asynchronous event-driven SNNs operate directly on raw event streams, making them a promising paradigm for achieving low-latency and low-power processing on neuromorphic hardware. However, current event-driven SNNs are explored mainly for classification tasks and have only rarely been applied to object detection. Moreover, their energy efficiency is typically assessed through simulation rather than measured on actual neuromorphic hardware. Future hardware deployments will be essential for obtaining real power consumption. In parallel, exploring energy-efficient event-based detectors on conventional CPUs or GPUs is also a promising direction. This can be achieved by designing lightweight object detection models or developing asynchronous event-driven architectures. Besides, exploring quantized ANNs for event-based vision tasks may also be a viable low-latency solution.

\begin{table*}[t]
   \caption{Comparison of state-of-the-art event-based object detectors on the 1Mpx Detection dataset~\cite{perot2020learning}.}
   \vspace{-0.15cm}
   \label{tab:1mpx_results}
   \scriptsize
   \begin{center}
   \renewcommand{\arraystretch}{1.10}
   \setlength{\tabcolsep}{1.40mm}{
   \begin{tabular}{l cc cccc ccccc}
        \toprule
        Method  & Type & Asynchronous & Representation & Backbone & Head  & Temporal & \# Param. (M) & mAP$_{50}$ & mAP$_{50:95}$    \\
        \hline
        RED~\cite{perot2020learning} & CNN-based & No & Voxel grid & CNN+RNN & SSD & ConvLSTM &  24.1 & - & 0.430 \\
        ASTMNet~\cite{li2021asynchronous} & CNN-based & No & Event embedding & CNN+RNN & SSD & Rec-Conv & 39.6 & - & 0.483 \\
        TEDNet~\cite{yen2024tracking} & CNN-based & No & Voxel grid & CNN+RNN & CenterTrack & ConvGRU & 21.3 & 0.562 & 0.312 \\
        AED~\cite{liu2023motion} & CNN-based & No & TAF & CNN & YOLOX & No & 14.8 & - & 0.344 \\
        Peng~\emph{et al.}~\cite{peng2023better} & CNN-based & No & Hyper histogram & CNN & YOLOv5 & No & - & - & 0.484 \\
        CED~\cite{zhu2024spatio} & CNN-based & No & DTR & CNN & CDE & LRC & 19.8 & - & 0.487 \\
        E-detector~\cite{zhang2023detector} & Transformer-based & No & Reconstructed image & Transformer & DETR & No & - & 0.467 & -\\
        RVT-B~\cite{gehrig2023recurrent} & Transformer-based & No & 4D tensor & Transformer+RNN & YOLOX & LSTM & 18.5 & 0.701 & 0.476 \\
        LEOD-RVT-S~\cite{wu2024leod} & Transformer-based & No & 4D tensor & Transformer & YOLOX & Time-flip TTA & 9.9 & - & 0.467 \\
        S5-ViT-B~\cite{zubic2024state} & Transformer-based & No & 4D Tensor & Transformer+SSM & YOLOX & SSM & 17.5 & - & 0.472 \\
        Zubi{\'c}~\emph{et al.}~\cite{zubic2023chaos} & Transformer-based & No & ERGO-12 & Transformer & YOLOV6 & No & - & - & 0.406 \\
        GET~\cite{peng2023get} & Transformer-based & No & Group token & Transformer+RNN & YOLOX & ConvLSTM & 21.9 & - & 0.484 \\
        EvRT-DETR-B~\cite{torbunov2025evrt} & Transformer-based & No & 4D tensor & Transformer+RNN & DETR &  RNN & 57.1 & - & \textbf{0.501} \\
        SMamba~\cite{yang2025smamba} & SSM-based & No &  Voxel grid & SSM+RNN & YOLOX &  STCA & 16.1 & - & 0.493\\
        \hline
        EGSST-E~\cite{wu2024egsst} & GNN-based & Yes & Graph & GNN+LinearViT & YOLOX & Yes & 12.3 & - & \textbf{0.502} \\
        \hline
        Ahmed~\cite{ahmed2024hybrid} & Hybrid SNN-ANN & No & 4D tensor & SNN & YOLOX & Yes & 6.6 & - & 0.270 \\
        EAS-SNN~\cite{wang2024eas} & SNN-based & No & ARSNN & SNN & YOLOX & Yes & 25.3 & 0.553 & 0.363 \\
        Li~\emph{et al.}~\cite{li2026spike} (M) & SNN-based & No & SDA-MTF & SNN & YOLOX & Yes & 25.3 & 0.591 & \textbf{0.304} \\
        SpikingViT-B~\cite{yu2024spikingvit} & SNN-based & No & TEE & SNN & YOLOX & Yes & 21.5 & 0.616 & 0.394 \\
        \bottomrule
    \end{tabular}}
\end{center}
\noindent These bounded categories refer to image-like ANN-based models (i.e., CNNs, Transformers, and SSMs), asynchronous GNN-based models, and energy-efficient SNN-based models.
\vspace{-0.20cm}
\end{table*}

\begin{table*}[htbp]
  \caption{Comparison of state-of-the-art object detectors on the PKU-DAVIS-SOD dataset~\cite{li2023sodformer} and the DSEC-Detection dataset~\cite{gehrig2022pushing}.}
  \vspace{-0.25cm}
  \label{tab:multimodal_methods}
  \scriptsize
  \begin{center}
    \renewcommand{\arraystretch}{1.10}
    \setlength{\tabcolsep}{0.55mm}{
    \begin{tabular}{l ccccccc cc cc}
    \toprule
    \multirow{2}*{Modality} & \multirow{2}*{Method} & \multirow{2}*{Representation}  & \multirow{2}*{Backbone} & \multirow{2}*{Head} & \multirow{2}*{Temporal} & \multirow{2}*{Fusion mode} & \multirow{2}*{\# Param. (M)} & \multicolumn{2}{c}{PKU-DAVIS-SOD} & \multicolumn{2}{c}{DSEC-Detection} \\ \cline{9-10} \cline{11-12} & & & & & & & & mAP$_{50}$ & mAP$_{50:95}$ & mAP$_{50}$ & mAP$_{50:95}$ \\
    \hline
    \multirow{5}*{Frame} & Faster R-CNN~\cite{ren2016faster} & RGB frame & CNN & R-CNN & No & - & 41.9 & 0.443 & - & 0.523 & 0.358 \\
    & YOLOX~\cite{przewlocka2024poweryolo} & RGB frame & CNN & YOLOX & No & - & 16.5 & 0.509 & 0.274 & 0.578 & 0.385 \\
    & Swin Transformer~\cite{liu2021swin} & RGB frame & Transformer & R-CNN & No  & - & 15.8 & 0.523 & 0.277 & 0.520 & 0.341 \\
    & SODFormer$^\ast$~\cite{li2023sodformer} & RGB frame & Transformer & DETR & TTE & - & 35.8 & 0.489	& 0.195 & 0.641 & 0.428 \\
    & HDI-Former$^\ast$~\cite{li2026rethinking} & RGB frame & Transformer & R-CNN & No  & - & 44.8 & 0.486 & 0.193 & 0.648	& 0.445 \\
    \hline
    \multirow{5}*{Event} & AED~\cite{liu2023motion} & TAF & CNN & YOLOX & No & - & 14.8 & 0.457 & 0.230 & 0.432 & 0.271 \\
    & RVT-B~\cite{gehrig2023recurrent} & 4D tensor & Transformer+RNN & YOLOX & LSTM & - & 18.5 & 0.503 & 0.256 & 0.442 & 0.277 \\
    & S5-ViT~\cite{zubic2024state} & 4D tensor & Transformer+SSM & YOLOX  & SSM & - & 17.5 & 0.466 & 0.232 & 0.387 & 0.238 \\
    & SODFormer$^\lozenge$ & Event image & Transformer & DETR & TTE & - & 35.8 & 0.334 & 0.128 & 0.409 & 0.266 \\
    & HDI-Former$^\lozenge$ & Event image & Spiking Transformer & R-CNN & SNN & - & 20.3 & 0.353 & 0.157 & 0.423 & 0.279 \\
    \hline
    \multirow{5}*{Event+Frame} & RENet~\cite{zhou2023rgb} & Event image & CNN & MOC & E-TMA & Bi-directional fusion & 59.8 & 0.549 & 0.288 & 0.490 & 0.316 \\
    & DAGr-ResNet50~\cite{gehrig2024low} & Graph & CNN+GNN & YOLOX & Graph & Concatenation & 34.6 & - & - & 0.660  & 0.419 \\
    & SODFormer~\cite{li2023sodformer} & Voxel grid & Transformer & DETR & TTE & Asynchronous fusion & 84.8 & 0.504 & 0.207 & 0.676 & 0.452 \\
    & FAOD~\cite{zhang2024frequency} & 4D tensor & CNN+RNN & YOLOX & LSTM & Align module & 20.3 & 0.575 & 0.305 & 0.635 & 0.425 \\
    & HDI-Former~\cite{li2026rethinking} & Event image & Transformer & Mask-RCNN & SNN & Dynamic interaction & 64.2 & - & - & 0.691 & 0.467 \\
    \bottomrule
    \end{tabular}}
  \end{center}
  \noindent Three bounded categories denote two single-modal object detection methods using either frames or events, and multimodal approaches that combine both modalities.
  \vspace{-0.30cm}
\end{table*}

\section{Benchmarking and Empirical Analysis}\label{sec:experiment}
This section begins with a systematic benchmarking of neuromorphic object detectors, providing quantitative comparisons. Then, we perform an attribute-based study to better understand the benefits and limitations of object detectors.

\subsection{Performance Benchmarking}\label{sec:performance}
In this part, we will conduct in-depth experimental analysis from both single-modal and multimodal perspectives.

\textbf{Comparison of Single-modal Methods.} As shown in Table~\ref{tab:gen1_results} and Table~\ref{tab:1mpx_results}, we present a comprehensive experimental survey comparing of state-of-the-art single-modal event-based object detectors on the Gen1 Detection dataset~\cite{de2020large} and the 1Mpx Detection dataset~\cite{perot2020learning}. From Table~\ref{tab:gen1_results}, Transformer-based methods currently achieve good performance, while EvRT-DETR-B~\cite{torbunov2025evrt} achieves the best performance on the Gen1 dataset with an mAP of 0.527. GNN-based methods offer low-latency asynchronous processing but suffer from limited detection accuracy. SNN-based methods aim to design energy-efficient processing that simultaneously approximates the corresponding ANN architecture. According to Table~\ref{tab:1mpx_results}, due to the high resolution and large scale of the 1Mpx Detection dataset, most current testing methods are CNN-based and Transformer-based. Note that, EvRT-DETR-B~\cite{torbunov2025evrt}, integrating Transformer and RNN architectures, has achieved the best detection performance on the 1Mpx Detection dataset~\cite{perot2020learning} with an mAP of 0.501. While single DVS events can achieve satisfactory results in ideal scenarios, achieving high object detection performance becomes notably challenging in scenarios like static or slow-motion situations that require fine textures.

\begin{figure*}[htbp]
\centering
\centerline{\includegraphics[width=\linewidth]{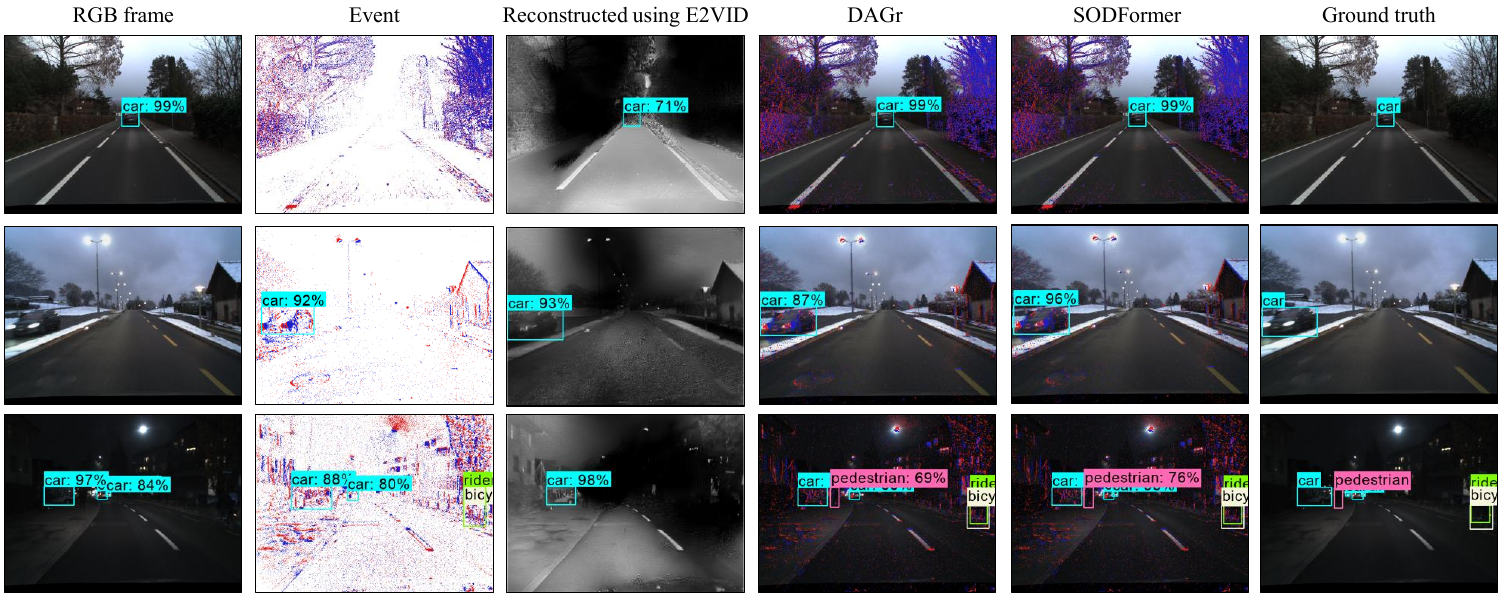}}
\caption{Representative visualizations of different object detection results on the DSEC-Detection dataset~\cite{gehrig2022pushing}. A single-modal model struggles to detect objects in low-light conditions. However, the joint frameworks can achieve robust object detection in challenging scenarios by leveraging both events and frames.}
\label{fig:multimodal_results}
\vspace{-0.30cm}
\end{figure*}

\textbf{Comparison with Multimodal Methods.} As shown in Table~\ref{tab:multimodal_methods}, we compare some state-of-the-art multimodal object detectors using events and frames on two widely used datasets (i.e., PKU-DAVIS-CAR~\cite{li2019event} and DSEC-Detection~\cite{gehrig2022pushing}). Note that, the joint frameworks improve consistency by integrating events and frames, outperforming single-modal methods. The FAOD~\cite{zhang2024frequency} achieves the best performance on the PKU-DAVIS-SOD dataset with mAP$_{50}$ and mAP$_{50:95}$ metrics of 0.575 and 0.305. Additionally, the HDI-Former~\cite{li2026rethinking} attains the highest performance on the DSEC-Detection dataset with mAP$_{50}$ and mAP$_{50:95}$ metrics of 0.691 and 0.467.

We further showcase representative visualization results from the DSEC-Detection dataset~\cite{gehrig2022pushing} in Fig.~\ref{fig:multimodal_results}. Note that, multimodal object detectors outperform the single-modal methods, including events, RGB frames, and reconstruction images using E2VID~\cite{rebecq2019events}. The joint detection frameworks excel in challenging conditions (e.g., motion blur and low light) by leveraging the high temporal resolution and high dynamic range inherited from DVS events, complemented by the fine textures provided by RGB frames. We can find that RGB frames fail to detect objects in low-light scenarios.

\subsection{Attribute-based Study}\label{sec:attribute}
This section explores how factors influence the detection performance. By studying these attributes, we aim to uncover the strengths of these models while identifying key challenges.

\begin{table}[t]
   \caption{Comparison of a typical event-based object detector~\cite{zubic2023chaos} using various input event representations on two datasets.}
   \label{tab:event_representation_results}
   \scriptsize
    \begin{center}
        \renewcommand{\arraystretch}{1.10}
        \setlength{\tabcolsep}{3.30mm}{
            \begin{tabular}{l ccc}
                \toprule
                \multirow{2}*{Event representation} & \multirow{2}*{\# Channels} & \multicolumn{2}{c}{mAP$_{50:95}$} \\ \cline{3-4} & & Gen1~\cite{de2020large} & 1Mpx~\cite{perot2020learning} \\
                \hline
                Binary image~\cite{rebecq2017real} & 2 & 0.339 & 0.327 \\
                Event count image~\cite{maqueda2018event} & 2 & 0.343 & 0.330  \\
                Reconstructed image~\cite{rebecq2019events} & 1 & 0.496 & 0.401 \\
                Voxel grid~\cite{zhu2019unsupervised} & 12 & 0.395 & 0.375 \\
                Time surface~\cite{lagorce2016hots} & 12 & 0.490 & 0.383 \\
                EST~\cite{gehrig2019end} & 12 & 0.485 & 0.382 \\
                TROE volume~\cite{baldwin2022time} & 12 & 0.436 & 0.381 \\
                ERGO~\cite{zubic2023chaos} & 12 & \textbf{0.504} & \textbf{0.406} \\
                \bottomrule
        \end{tabular}}
    \end{center}
    \vspace{-0.30cm}
\end{table}

\textbf{Analysis of Input Event Representations.} As depicted in Table~\ref{tab:event_representation_results}, we compare various input event representations for a representative open-source event-based object detector~\cite{zubic2023chaos} on the Gen1 Detection dataset~\cite{de2020large} and the 1Mpx Detection dataset~\cite{perot2020learning}. For event representations with adjustable channel numbers (e.g., event count image~\cite{maqueda2018event}, voxel grid~\cite{zhu2019unsupervised}, and ERGO~\cite{zubic2023chaos}), we set the number of channels to 12. Note that, Zubi{\'c} ~\emph{et al.}~\cite{zubic2023chaos} propose ERGO, an end-to-end framework that learns event representations jointly with the object detection backbone, achieving state-of-the-art performance. The results demonstrate that end-to-end learned representations consistently outperform event images and handcrafted features for event-based object detection.

\begin{table}
   \caption{The influence of temporal aggregation size in temporal modeling modules on the Gen1 Detection dataset~\cite{de2020large}.}
   \label{tab:temporal_modeling_results}
   \scriptsize
    \begin{center}
        \renewcommand{\arraystretch}{1.10}
        \setlength{\tabcolsep}{2.00mm}{
            \begin{tabular}{c cccc}
                \toprule
                \multirow{2}*{Temporal size $T$} & \multicolumn{2}{c}{Rec-Conv~\cite{li2021asynchronous}} & \multicolumn{2}{c}{TAM~\cite{guo2024spatio}} \\ \cline{2-3} \cline{4-5} & mAP$_{50:95}$ & Runtime (ms) & mAP$_{50:95}$ & Runtime (ms) \\
                \hline
                1 & 0.386 & 28.62 & 0.444 & 28.4 \\
                3 & 0.467 & 35.61 & 0.499 & 32.5 \\
                6 & 0.483 & 90.35 & 0.498 & 41.8 \\
                9 & 0.487 & 212.09 & 0.502 & 47.9 \\
                \bottomrule
        \end{tabular}}
    \end{center}
    \noindent Runtime refers to the time it takes to process an event bin and produce a prediction.
    \vspace{-0.30cm}
\end{table}

\begin{table}[t]
    \caption{The influence of multimodal fusion strategies for SODFormer~\cite{li2023sodformer} on the DSEC-Detection dataset~\cite{gehrig2022pushing}.}
    \label{tab:multimodal_fusion_results}
    \scriptsize
    \begin{center}
    \renewcommand{\arraystretch}{1.10}
    \setlength{\tabcolsep}{2.60mm}{
        \begin{tabular}{l ccc}
            \toprule
            Fusion strategy & Type & mAP$_{50:95}$ & Runtime ($ms$) \\
            \hline
            NMS & Synchronous & 0.250 & 14.8 \\
            Averaging & Synchronous & 0.258 & 71.7 \\
            Concatenation & Synchronous & 0.262 & 71.2 \\
            Asynchronous fusion~\cite{li2023sodformer} & Asynchronous & 0.271 &  72.2 \\
            \bottomrule
    \end{tabular}}
\end{center}
\noindent Runtime refers to the time it takes to process an event bin and produce a prediction.
\vspace{-0.30cm}
\end{table}

\textbf{Analysis of Temporal Modeling.} To analyze the impact of temporal aggregation size for event bins on detection performance, we compare various aggregation sizes on representative temporal modeling modules using the Gen1 Detection dataset (see Table~\ref{tab:temporal_modeling_results}). More precisely, we present two typical temporal modeling modules (i.e., Rec-Conv~\cite{li2021asynchronous} and TAM~\cite{guo2024spatio}) with different temporal aggregation sizes (i.e., 1, 3, 6, and 9) for comparison. As we increase the temporal aggregation size, the performance tends to improve consistently. This indicates that a larger temporal bin size captures more useful long-range temporal cues in the temporal modeling modules, leading to better detection performance. However, increasing the temporal aggregation size also raises computational speed. In some scenarios, such as detecting small objects~\cite{he2026towards} and recognizing objects under heavy noise or occlusion, temporal information plays an important role in improving detection accuracy. It's also important to note that the optimal temporal aggregation size is not fixed. It depends on factors like how fast the event camera or objects are moving. Thus, it is important to design velocity-invariant surfaces~\cite{goyal2023moveenet} that eliminate the need to adjust temporal aggregation sizes for different camera or object speeds.

\begin{figure}
    \centering
    \centerline{\includegraphics[width=\linewidth]{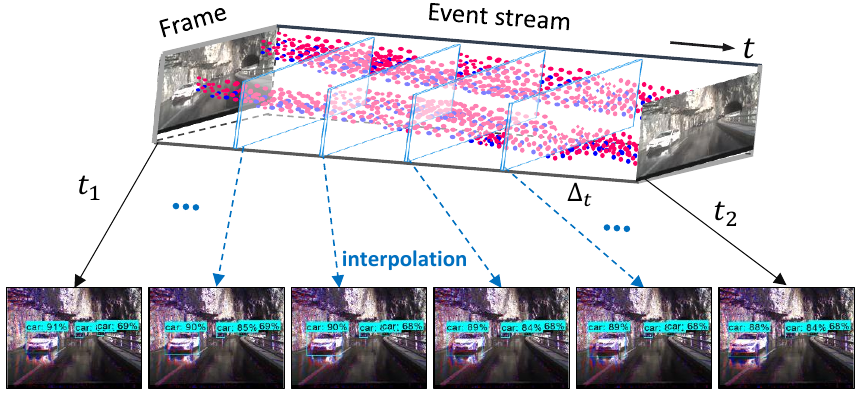}}
    \caption{Visualization of asynchronous inference using the asynchronous fusion strategy~\cite{li2026rethinking}. The top figure displays two adjacent frames and the event stream between, while the bottom figures show high-rate object detection results for the two frames and four event stream timestamps.}
    \label{fig:asynchronous_inference}
    \vspace{-0.30cm}
\end{figure}

\textbf{Analysis of Multimodal Fusion.} As shown in Table~\ref{tab:multimodal_fusion_results}, we compare some typical multimodal fusion strategies for a joint detection framework (i.e., SODFormer~\cite{li2023sodformer}) using frames and events on the DSEC-Detection dataset. Specifically, we select a post-processing operation (i.e., NMS~\cite{hosang2017learning}), two end-to-end feature aggregation fusion strategies (i.e., averaging and concatenation), and an asynchronous attention-based fusion module for comparison. Note that, the asynchronous fusion operation achieves the best performance while maintaining comparable inference times to other methods. In particular, the inference frequency of these synchronous fusion methods is limited by the RGB frame rate. In contrast, the asynchronous fusion method leverages complementary information from two modalities and overcomes the limited inference frequency imposed by the RGB frame rate. Furthermore, we illustrate an instance of an asynchronous fusion strategy for high-speed inference using HDIFormer~\cite{li2026rethinking} in Fig.~\ref{fig:asynchronous_inference}. We can observe that the four figures in the middle display the detection results of the event modality between two RGB frames. In other words, the asynchronous fusion strategy bridges the gap between two adjacent RGB frames, enabling high-frequency object detection in an asynchronous manner.

\textbf{Analysis of Energy-efficiency.} To evaluate the energy efficiency of SNN-based object detection models, we compare each SNN with its ANN counterpart using the same architecture (see Table~\ref{tab:energy_efficient_results}). The energy estimates are computed using theoretical AC and MAC operation counts under a standard hardware model (i.e., 32-bit floating-point operations implemented in 45 nm technology~\cite{horowitz20141}), where a MAC costs $E_M$ $=$4.6 pJ and an AC costs $E_A $$=$ 0.9 pJ. These results reflect simulated energy values rather than measurements from physical neuromorphic hardware. To avoid overinterpreting absolute energy numbers, we report the energy efficiency ratio between each SNN and its ANN baseline, which provides a consistent basis for comparison. As shown in Table~\ref{tab:energy_efficient_results}, SNN-based models consistently exhibit higher energy efficiency than their ANN counterparts with equivalent architectures. In particular, SNN-CenterNet~\cite{bodden2024spiking} and SpikeYOLO~\cite{luo2024integer} achieve efficiency ratios of  5.65$\times$ and 6.09$\times$, respectively. In addition, the firing rate is a key indicator for understanding the internal mechanisms of SNNs and optimizing network design~\cite{su2023deep}. In summary, combining SNN-based models with neuromorphic computing chips is expected to reduce power consumption while achieving performance comparable to ANN models. However, these conclusions are currently based solely on theoretical MAC and AC operation counts on specialized hardware. Experimental validation on actual neuromorphic platforms is needed to confirm these energy-efficiency benefits in real-world applications.

\begin{table}[t]
    \caption{Comparison of energy consumption with various SNN-based object detection models on the Gen1 Detection dataset~\cite{de2020large}.}
    \vspace{-0.30cm}
    \label{tab:energy_efficient_results}
    \scriptsize
    \begin{center}
    \renewcommand{\arraystretch}{1.10}
    \setlength{\tabcolsep}{0.50mm}{
        \begin{tabular}{l ccccc}
            \toprule
            Model & \# Param. (M) & Time step & Firing rate & Efficiency ratio & mAP$_{50:95}$ \\
            \hline
            SNN-MobileNet~\cite{cordone2022object} & 24.26 & 5 & 29.44\% & - & 0.147 \\
            SNN-DenseNet~\cite{cordone2022object} & 8.2 & 5 & 37.20\% & - & 0.189 \\
            MS-ResNet18~\cite{su2023deep} & 9.49 & 5 & 17.08\% & 2.43$\times$ & 0.285 \\
            Sew-ResNet18~\cite{su2023deep} & 9.56 & 5 & 18.80\% & 2.00$\times$ & 0.286 \\
            EMS-ResNet18~\cite{su2023deep} & 9.34 & 5 & 20.09\% & 4.91$\times$ & 0.286 \\
            FP-DAGNet~\cite{zhang2024automotive} & 22.00 & 3 & 19.10\% & - & 0.223 \\
            SFOD~\cite{fan2024sfod} & 11.9 & - & 24.04\% & - & 0.321 \\
            SNN-CenterNet~\cite{bodden2024spiking} & 12.97 & 5 & 17.40\% & 5.65$\times$ & 0.229 \\
            Spiking-ViT-B~\cite{yu2024spikingvit} & 21.48 & C & 18.01\% & 2.28$\times$  & 0.394 \\
            SpikeYOLO~\cite{luo2024integer} & 23.1 & 4 & - & \textbf{6.09$\times$} & \textbf{0.404} \\
            \bottomrule
    \end{tabular}}
\end{center}
\vspace{-0.30cm}
\end{table}

\section{Discussions and Future Directions}\label{sec:discussion}
This section will discuss several open issues and highlight active research directions for neuromorphic object detection.

\subsection{New Evaluation Metrics for Low-latency Detection}

Effective metrics are vital for advancing the neuromorphic object detection community, as they offer guidance for setting benchmarks, evaluating, and enhancing new methodologies. However, current metrics (e.g., mAP) for event-based object detection are largely inherited from the frame-based domain (see Table~\ref{tab:object_detection_metric}) and fail to capture the unique attributes of event cameras. A key advantage of event cameras is their microsecond-level temporal resolution, which enables low-latency detection compared to conventional cameras. Standard metrics like mAP treat detections at different timestamps equally without accounting for temporal latency, making them suboptimal for evaluating object detectors in low-latency applications. In safety-critical autonomous driving scenarios, detecting objects like vehicles or pedestrians with minimal delay is crucial. A more illustrative metric may be detection delay, defined as the temporal gap between when an object first triggers events and when it is detected by the model. Integrating temporal latency with detection accuracy in new metrics (e.g., time-to-first-detection) can better reflect the unique advantages of neuromorphic sensors in low-latency scenarios. Therefore, there is a clear need to develop novel evaluation metrics that jointly account for detection accuracy, temporal latency, and the sequential nature of event streams~\cite{mao2019delay, sobti2021vmap}. Moreover, future benchmarking of neuromorphic object detection systems should also report the EDP for the trade-off between energy consumption and latency, as it provides a more comprehensive evaluation of system-level performance.

\subsection{Large-scale Multimodal Multi-task Datasets}
Most existing neuromorphic object detection datasets (see Table~\ref{tab:object_detection_dataset}) are limited in data annotation scale, typically designed for just one specific task, and often involve re-annotating the same data repeatedly (e.g., DSEC~\cite{gehrig2021dsec}). Clearly, the neuromorphic vision domain lacks the large-scale multimodal datasets and benchmarks that are common in the frame-based domain. In fact, there is a need for a comprehensive and large-scale labeled dataset that is well-established and long-lasting. A standardized dataset should support multiple modalities (e.g., event, RGB, infrared, and LiDAR) and various tasks (e.g., object detection, object tracking, and stereo vision~\cite{li2026towards}). Both simulated datasets and real-world challenging datasets are worth exploring. On one hand, physics-based event camera simulators~\cite{rebecq2018esim, han2024physical} integrated with unreal engine systems allow for the creation of highly realistic datasets, ideal for training large multimodal models. These 3D simulation environments allow the generation of high-fidelity event data within dynamic interactive settings, providing a platform for developing and evaluating closed-loop perception–action systems in embodied neuromorphic agents. On the other hand, multi-sensor prototype platforms should be developed for autonomous vehicles~\cite{iliasov2025multichannel} or agile robots to collect multimodal data and perform multi-task annotations. In particular, these datasets should include high-speed motion scenarios to fully exploit the microsecond-level temporal resolution of neuromorphic cameras.

\subsection{Low-latency Object Detection for Agile Robots}
Low-latency object detection maximizes the advantages of neuromorphic cameras in real-world high-speed scenes~\cite{durr2026outplaying}, making them ideal for agile robots such as autonomous racing cars, high-speed drones, and robotic dogs. Ongoing optimization of detection methods aims to boost detection speed while maintaining accuracy, including synchronous lightweight neural networks and asynchronous event-based strategies. One solution involves designing lightweight networks and applying model compression techniques to address the trade-off between accuracy and speed. However, this approach still converts events into synchronous image-like formats, creating a bottleneck in detection inference speed. Asynchronous event-by-event processing paradigms~\cite{gehrig2024low, ieng2014asynchronous, tapia2024efft} are seen as a promising solution for high-speed applications, as they leverage the benefits of sparse data and low latency. Current asynchronous event-by-event processing methods like sparse convolution~\cite{messikommer2020event} and GNNs~\cite{li2021graph, verma2026event} enable low-latency object detection within milliseconds (see Section~\ref{sec:asynchronous}), but their accuracy remains inferior to synchronous methods using image-like representations. What's more, further validation of these object detection methods on asynchronous event-based circuits will be crucial moving forward.

\begin{figure}[t]
\centering
\includegraphics[width=\linewidth]{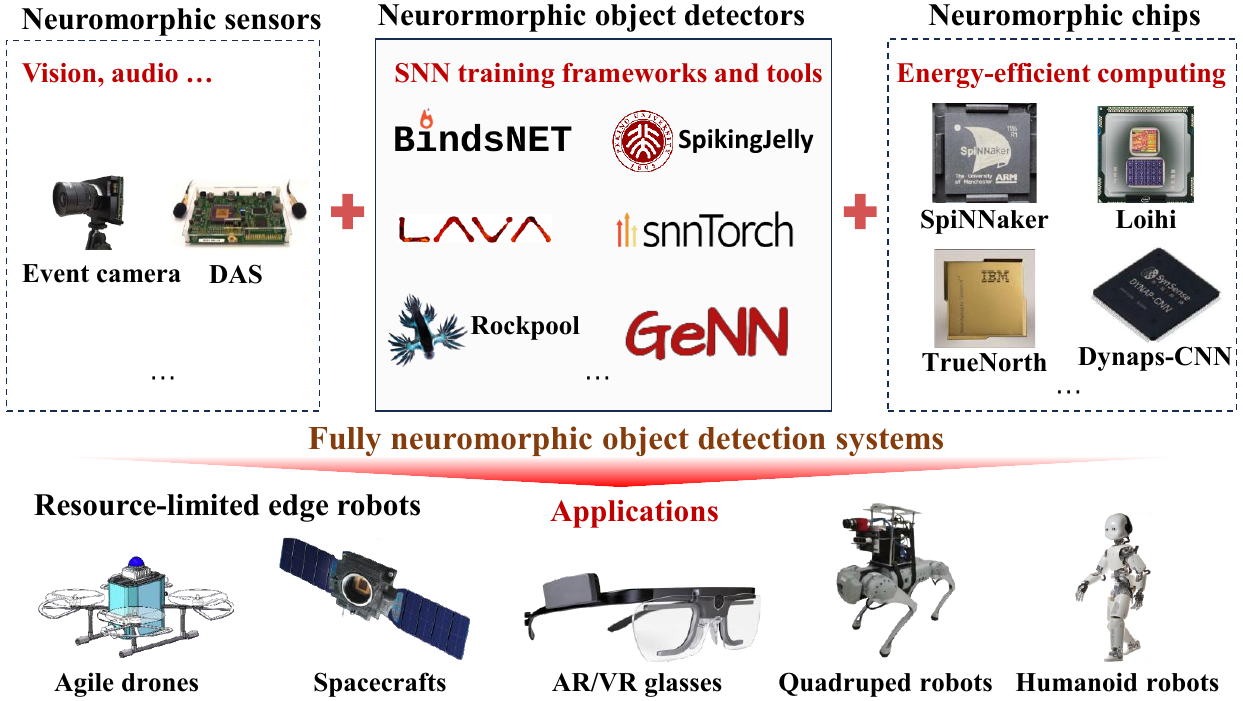}
\caption{A paradigm for fully neuromorphic object detection. This pipeline illustrates a complete system: neuromorphic sensors capture event streams, SNN-based algorithms process the data, and neuromorphic computing chips execute the energy-efficient computations. Representative frameworks are selected based on their SNN training support and active maintenance. Representative neuromorphic chips are chosen when commercially available and actively used in the neuromorphic community. Such systems offer substantial advantages in energy efficiency and latency, making them well-suited for resource-constrained edge platforms, such as agile drones and spacecraft.}
\label{fig:fully_neuromorphic}
\end{figure}

\subsection{Large Neuromorphic Multimodal Models}
Multimodal pre-trained models~\cite{kuo2022beyond} and large language-vision models (LLVMs)~\cite{wang2025towards} have shown significant performance gains in object detection tasks. However, these cutting-edge techniques have yet to be fully explored for neuromorphic vision, especially for object detection. In practice, developing pre-trained models or LLVMs specifically for neuromorphic object detection could offer substantial benefits. By training these models on large-scale multimodal datasets~\cite{yang2023event, klenk2024masked}, it is possible to build a comprehensive understanding of complex environments. Furthermore, integrating LLVMs into neuromorphic vision (e.g., EventGPT~\cite{liu2024eventgpt} and EventFlash~\cite{liu2026eventflash}) could facilitate the development of models that are not only capable of detecting objects but also reasoning about the semantic context and relationships between them~\cite{kudithipudi2025neuromorphic}. For example, LLVMs could enable object detection systems to infer high-level semantic cues from event streams, enhancing performance in real-world complex environments. Despite these promising opportunities for large neuromorphic multimodal models, several key challenges persist. The asynchronous nature of event data needs new architectures to effectively process this input format. Pre-trained models for neuromorphic vision tasks should take into account the unique properties of event streams, such as temporal cues and sparsity. Besides, optimizing the computational efficiency of LLVMs is essential to meet the low-latency and low-power demands of neuromorphic computing systems, particularly for resource-limited edge devices. 

\subsection{Fully Neuromorphic Object Detection Systems}
Fully neuromorphic systems~\cite{yao2024spike, paredes2024fully, jiang2025fully, zhang2020system, ma2022neuromorphic}, integrating event-based sensing and neuromorphic computing, have shown substantial benefits in both energy efficiency and low latency for various tasks. These systems usually use neuromorphic sensors to capture event streams, process them with SNN-based algorithms, and perform computations on neuromorphic computing chips~\cite{shrestha2022survey, zhou2023computational, zhang2026compute, theilman2025solving}. The development of fully neuromorphic object detection systems offers rapid object localization capabilities while maintaining high energy efficiency, making them particularly suitable for power-constrained edge platforms such as agile drones, autonomous vehicles, augmented reality, and space satellites (see Fig.~\ref{fig:fully_neuromorphic}). However, the current evaluation of SNN-based object detectors often relies on power consumption metrics obtained through simulations for specific neuromorphic chips, which may not fully represent the practical power consumption of these systems in real-world applications. To address this gap, it is essential to validate SNN-based object detection systems on actual neuromorphic computing platforms used in real-world scenarios. Furthermore, it is important to consider the design of hybrid ANN-SNN architectures that combine the strengths of both event-based and frame-based approaches, thereby enhancing overall performance in diverse and challenging scenarios. By addressing these challenges, fully neuromorphic systems could unlock new potentials for energy-efficient intelligent computing in power-constrained edge devices.

\subsection{Multi-agent Collaborative Object Detection}
High-speed objects, due to their rapid movements and extensive motion range, can quickly exit the narrow field of view of a single event camera. Collaborative object detection using multi-agent systems~\cite{qu2023elastic} presents a promising solution by expanding the field of view through the integration of perspectives from multiple cameras. This novel multi-view collaborative approach not only aims to reduce false alarms commonly associated with single-camera detection but also enhances the overall accuracy and robustness of object detection systems, particularly for high-speed moving objects or occlusion situations. Despite its advantages, several challenges persist in the realm of collaborative object detection. One major issue is ensuring spatiotemporal synchronization among the various neuromorphic cameras, as misalignment can lead to discrepancies and hinder effective fusion. Asynchronous fusion of multiple streams is another challenge, requiring advanced algorithms to seamlessly integrate event data captured from different perspectives. Furthermore, the computational complexity of processing large volumes of event data from multiple cameras can strain resources, highlighting the need for efficient algorithms that manage high-dimensional inputs while maintaining real-time performance. Ultimately, addressing these challenges is crucial for realizing the full potential of collaborative object detection in dynamic environments.

\begin{figure}[t]
    \centering
    \centerline{\includegraphics[width=0.95\linewidth]{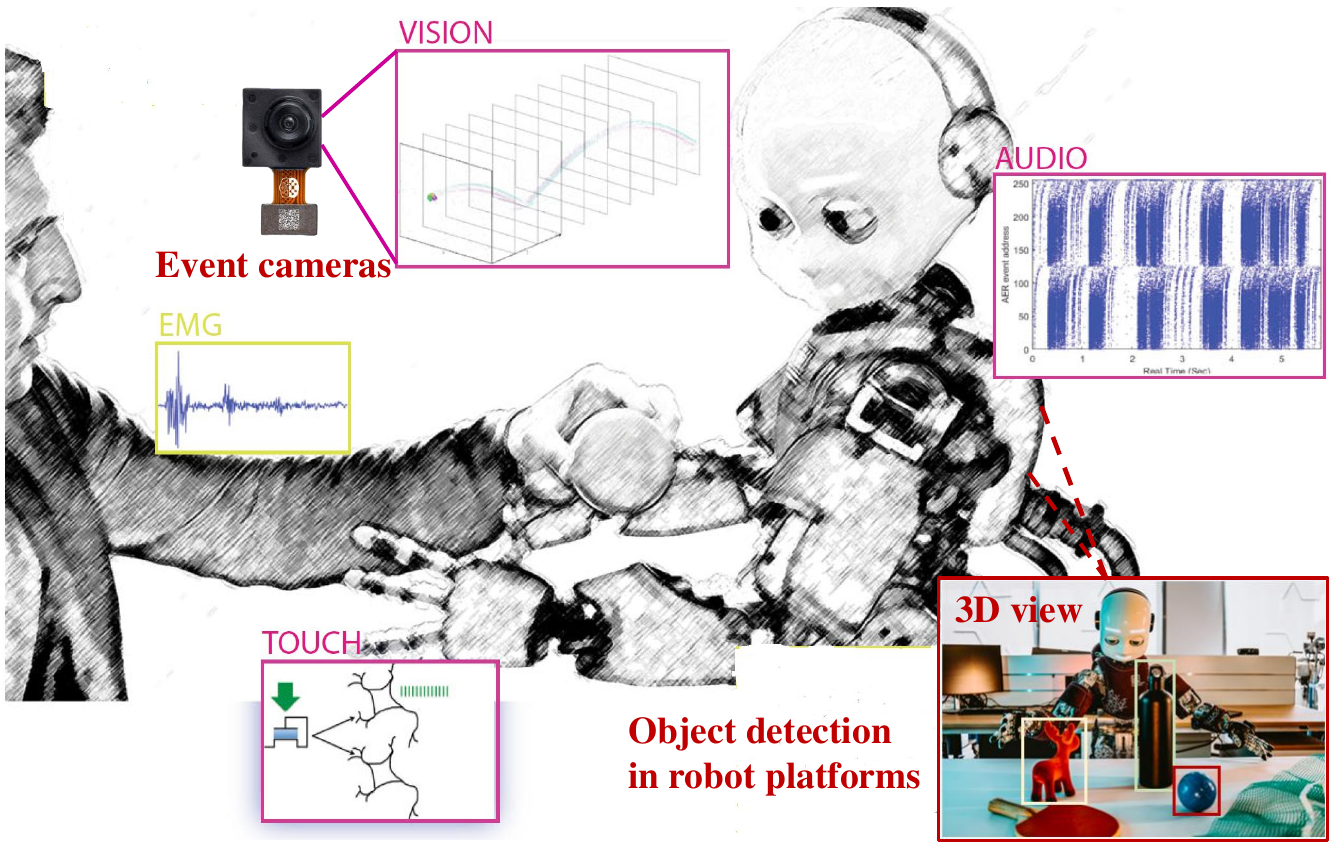}}
    \caption{The neuromorphic object detection system is integrated into the embodied iCub humanoid robot, enabling continuous learning and adaptation in complex and dynamic environments. The iCub robot (image ©IIT~\cite{bartolozzi2022embodied}) serves as a platform that utilizes neuromorphic sensors, including event cameras, dynamic audio sensors (DAS), and tactile sensors~\cite{bartolozzi2016robots, bartolozzi2018neuromorphic}.}
    \label{fig:embodied_neuromorphic}
\end{figure}

\subsection{Towards Embodied Neuromorphic Detection}
Neuromorphic cameras~\cite{posch2014retinomorphic} are originally designed to enhance machine vision for agile robotics, where perception operates in tight coupling with action. However, most current neuromorphic object detectors remain focused on single-task and offline evaluation. This limitation restricts their deployment on robotic platforms that require real-world interaction capabilities such as continuous object detection, planning, and control. A meaningful breakthrough in neuromorphic object detection requires integrating these detectors into embodied robot platforms, enabling closed-loop perception–action interactions and continuous adaptation in complex dynamic environments. For instance, embodied neuromorphic humanoid robots (e.g., iCub)~\cite{chicca2014neuromorphic, glover2016event, natale2017icub, bartolozzi2022embodied, d2025event, hajizada2022interactive} can leverage real-time visual feedback for various tasks such as rapid object detection for reaching and object shape estimation for grasping (see Fig.~\ref{fig:embodied_neuromorphic}). This integration allows agile robots to adaptively respond to environmental changes, facilitating robust multi-task performance and continuous learning in real-world scenarios. Nevertheless, advancing toward embodied neuromorphic detection faces several challenges~\cite{sandamirskaya2022neuromorphic, d2026benchmarking}: (i) How can we establish comprehensive event-based datasets that capture the complex dynamics of agent–object interactions in challenging scenarios, particularly under high-speed motions and varying lighting conditions? (ii) How should we develop unified frameworks to assess multi-task performance within complete perception–action cycles? (iii) How can we create fully neuromorphic robot systems that combine event-based sensing with neuromorphic computing to enable low-latency and energy-efficient object detection? Addressing these challenges is essential for advancing neuromorphic object detection from algorithms to real-world embodied intelligence.


\section{Conclusion}\label{conclusion}
To the best of our knowledge, this paper is the first comprehensive survey and benchmark of object detection using neuromorphic cameras. We begin by describing the problem, highlighting key challenges, comparing existing datasets, and rethinking current evaluation metrics. We then review existing neuromorphic object detectors from various research perspectives, including event representations, temporal modeling, multimodal fusion, asynchronous processing, low-latency processing, and energy-efficient computing. Finally, we provide a systematic benchmark of state-of-the-art neuromorphic object detectors and conduct an attribute-based analysis to better understand the strengths of current strategies. Several unresolved issues and research opportunities remain, particularly in achieving low power consumption and low-latency processing. We hope this in-depth survey serves as an effective way to understand the current state-of-the-art and encourages further innovative exploration in the neuromorphic community.

\section*{Acknowledgments}
This work was supported in part by the National Natural Science Foundation of China under Contracts (62425101, 62332002, and 62506190), the Fundamental and Interdisciplinary Disciplines Breakthrough Plan of the Ministry of Education of China under Contract (JYB2025XDXM108), the Beijing Science and Technology Plan under Contract (Z241100004224011), and the Shenzhen Science and Technology Program under Contract (KQTD20240729102051063). We sincerely thank Prof. Tobi Delbrück for his valuable suggestions on the revision of this paper. We also extend our gratitude to Dr. Dongyang Ma and Dr. Xu Liu for their contributions to this work.

\bibliographystyle{IEEEtran}
\bibliography{IEEEabrv,proc_ieee_references}

\vspace{-0.70cm}
\begin{IEEEbiography}[{\includegraphics[width=1in,height=1.0in, clip,keepaspectratio]{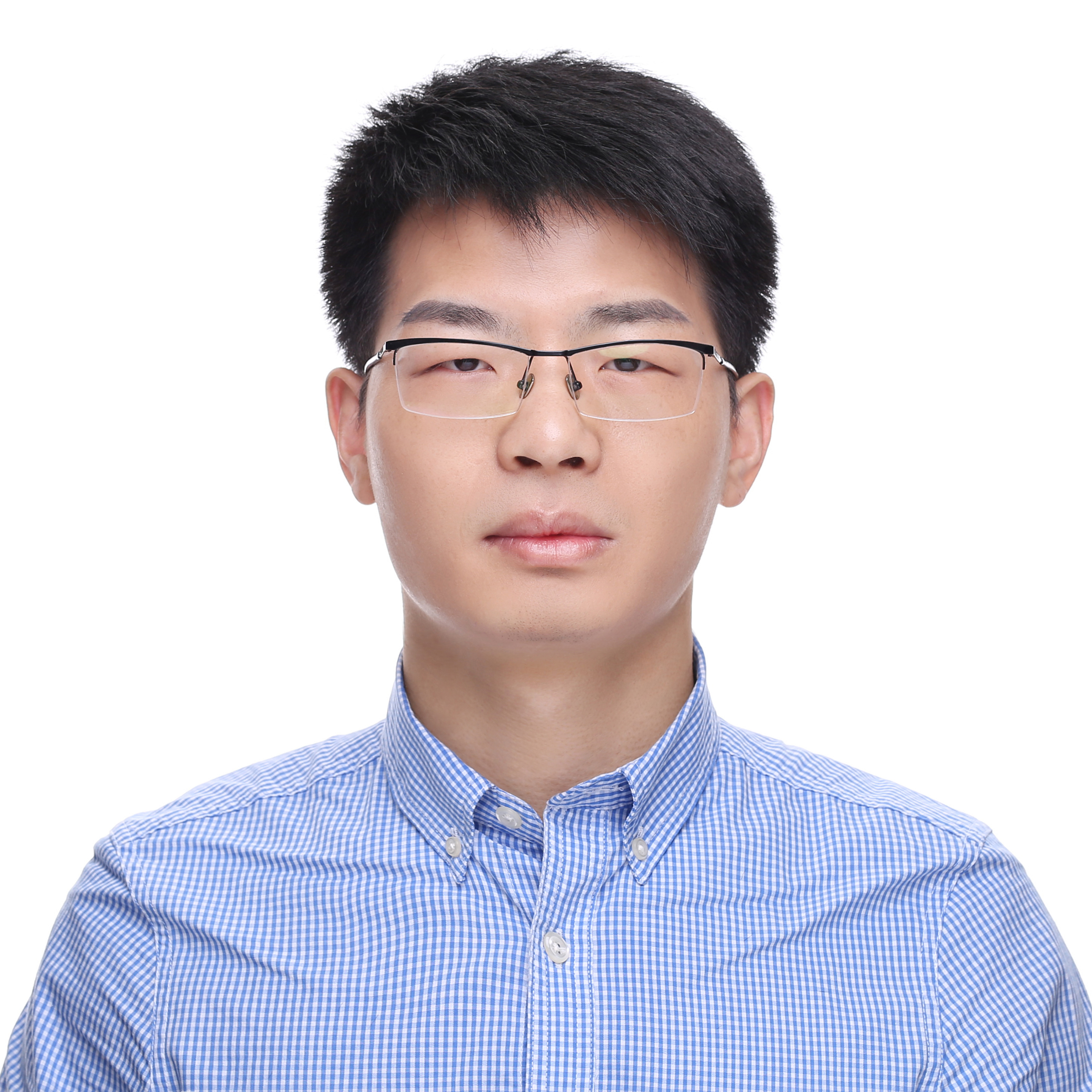}}]{Jianing Li} (IEEE Member) received the Ph.D. degree from the National Engineering Research Center for Visual Technology, School of Computer Science, Peking University, Beijing, China, in 2022. He is currently an associate researcher in the School of Computer Science at Peking University, Beijing, China. He is the author or coauthor of over 50 technical papers in refereed journals and conferences, such as TPAMI, TIP, CVRP, ICCV, NeurIPS, and ICML. He received the Outstanding Research Award from Peking University in 2020. He was honored with the Outstanding PhD Thesis Award from the CIE in 2024. His research interests include neuromorphic computing, event-based vision, neuromorphic engineering, and robot learning.
\end{IEEEbiography}

\begin{IEEEbiography}[{\includegraphics[width=1in,height=1in, clip,keepaspectratio]{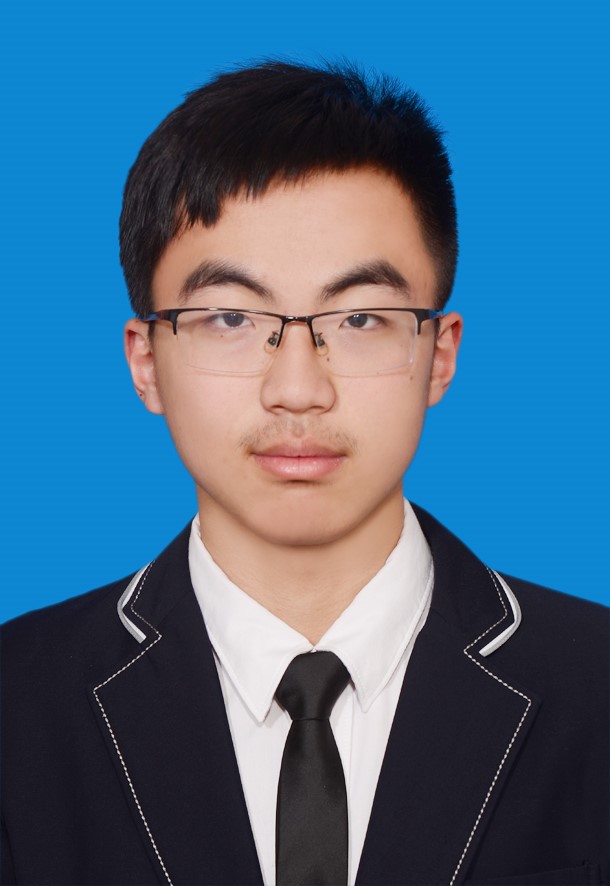}}]{Dianze Li} received the B.S degree from the School of Electronics Engineering and Computer Science, Peking University, Beijing, China, in 2022. He is currently pursuing a Ph.D. degree with the National Engineering Research Center for Visual Technology, School of Computer Science, Peking University, Beijing, China.
His current research interests include event-based vision, spiking neural networks, and neuromorphic engineering.
\end{IEEEbiography}

\begin{IEEEbiography}[{\includegraphics[width=1in, height=1in, clip, keepaspectratio]{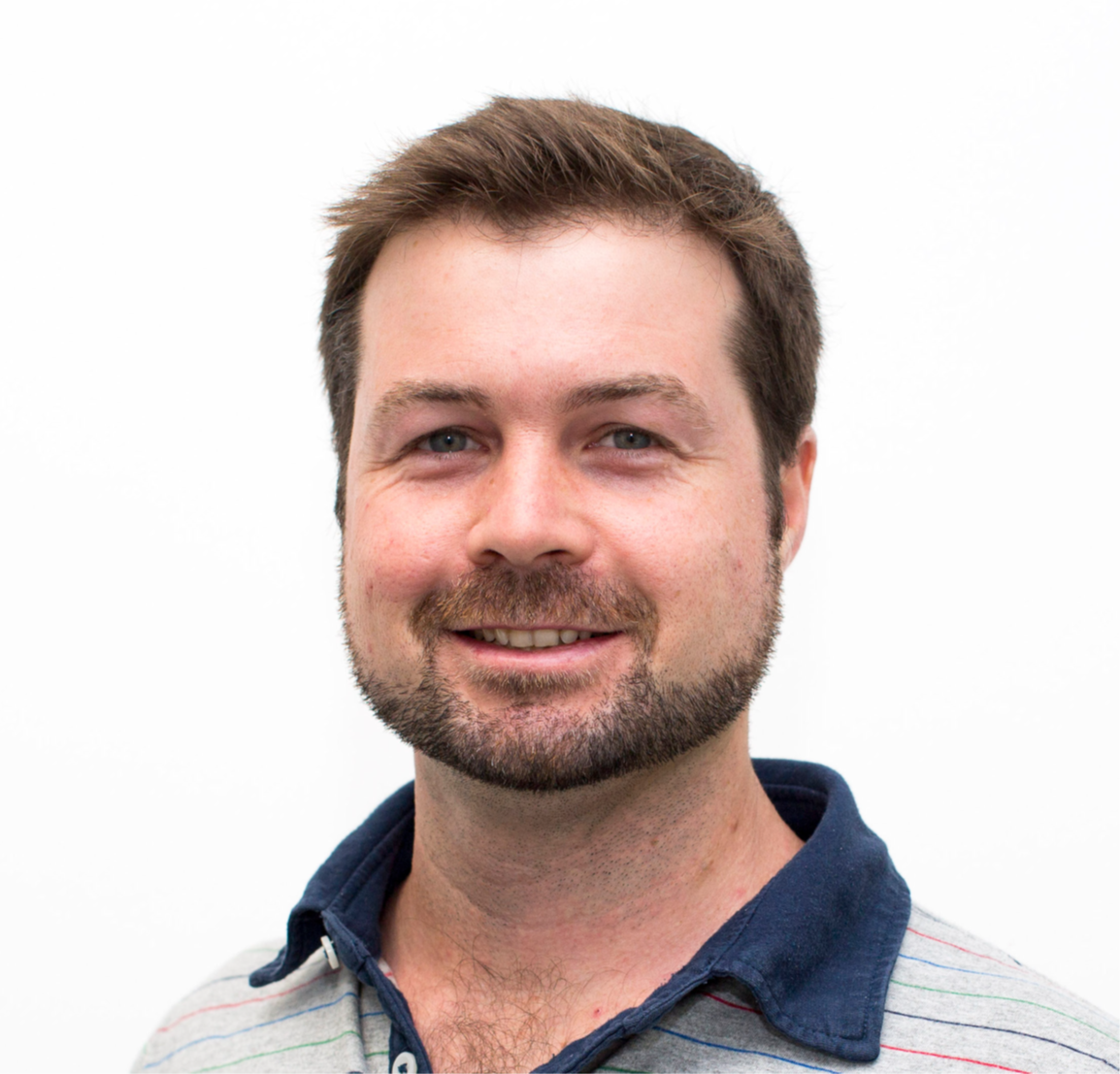}}] {Arren Glover} (IEEE Member) obtained a Bachelor of Mechatronic Engineering from the University of Queensland and a PhD in developmental robotics and on-line learning from the Queensland University of Technology, Australia. Since 2015 he has been with the Italian Institute of Technology’s Event-driven Perception for Robotics research line with the focus on developing real-time vision algorithms for the iCub robot using event-cameras. He is a senior researcher also leading industrial collaborations.
\end{IEEEbiography}

\begin{IEEEbiography}
[{\includegraphics[width=1in,height=1.25in,clip,keepaspectratio]{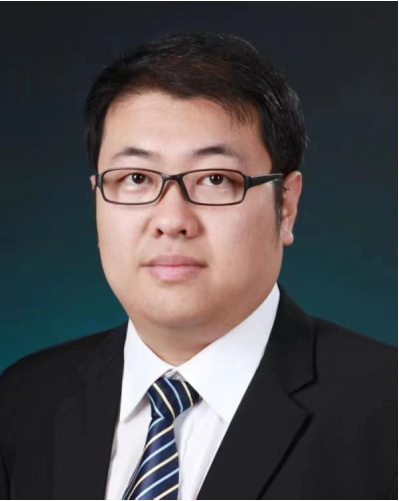}}]{Xiaopeng Fan}
(IEEE Senior Member) received the B.S. and M.S. degrees from the Harbin Institute of Technology (HIT), Harbin, China, in 2001 and 2003, respectively, and the Ph.D. degree from the Hong Kong University of Science and Technology, Hong Kong, in 2009. In 2009, he joined HIT, where he is currently a Professor. From 2003 to 2005, he was with Intel Corporation, China, as a Software Engineer. From 2011 to 2012, he was with Microsoft Research Asia, as a Visiting Researcher. From 2015 to 2016, he was with the Hong Kong University of Science and Technology, as a Research Assistant Professor. He has authored one book and more than 180 articles in refereed journals and conference proceedings. His research interests include video coding and transmission, image processing, and computer vision. He was the Program Chair of PCM2017, Chair of IEEE SGC2015, and Co-Chair of MCSN2015. He was an Associate Editor for IEEE 1857 Standard in 2012. He was the recipient of Outstanding Contributions to the Development of IEEE Standard 1857 by IEEE in 2013.
\end{IEEEbiography}

\vspace{-0.20cm}

\begin{IEEEbiography}[{\includegraphics[width=1.2in, height=1.2in, clip, keepaspectratio]{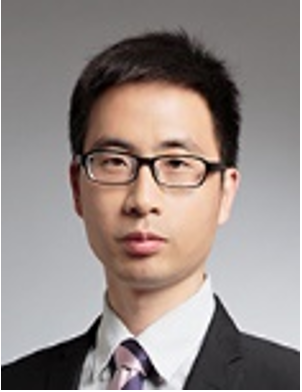}}] {Guoqi Li} received the Ph.D. degree from Nanyang Technological University, Singapore, in 2011. From 2011 to 2014, he was a Scientist with the Data Storage Institute and the Institute of High Performance Computing, Agency for Science, Technology and Research, Singapore. From 2014-2022, he was an Assistant professor and Associate professor at Tsinghua University, Beijing, China. Since 2022, he has been with the Institute of Automation, Chinese Academy of Sciences and the University of Chinese Academy of Sciences, where he is currently a Full professor. His current research interests include Brain-inspired Intelligence, Neuromorphic Computing, and Spiking Neural Networks. He has authored or co-authored more than 200 papers in a number of prestigious journals including Nature, Nature Communications, Science Robotics, Proceedings of the IEEE, and top AI conference such as ICLR, NeurIPS, ICML, and so on. 

Dr. Li has been actively involved in professional services such as serving as a Tutorial Chair, an International Technical Program Committee Member, a PC member, a Publication Chair, a Track Chair and workshop chair for several international conferences. He is an Editorial-Board Member for Control and Decision, and served as Associate Editors for IEEE Transactions on Neural Networks and Learning Systems, IEEE Transactions on Cognitive and Developmetal Systems, and Frontiers in Neuroscience: Neuromorphic Engineering. He is a reviewer for Mathematical Reviews published by the American Mathematical Society and serves as a reviewer for a number of prestigious international journals and top AI conferences including ICLR, NeurIPS, ICML, AAAI, and so on. He was the recipient of the 2018 First Class Prize in Science and Technology of the Chinese Institute of Command and Control, the Top ten scientific advances Award in China selected by the Ministry of science and technology, P.R. China as the backbone of the team member, and the 2020 Second Prize of Fujian Provincial Science and Technology Progress Award.  Dr. Li was honored with the Outstanding Young Talent Award from the Beijing Natural Science Foundation in 2021, and was selected to participate in the Hundred Talents Program of the Chinese Academy of Sciences in 2022. In 2023, Dr. Li was awarded the National Science Foundation for Distinguished Young Scholars of China.
\end{IEEEbiography}

\vspace{-0.20cm}

\begin{IEEEbiography}[{\includegraphics[width=1in, height=1in, clip, keepaspectratio]{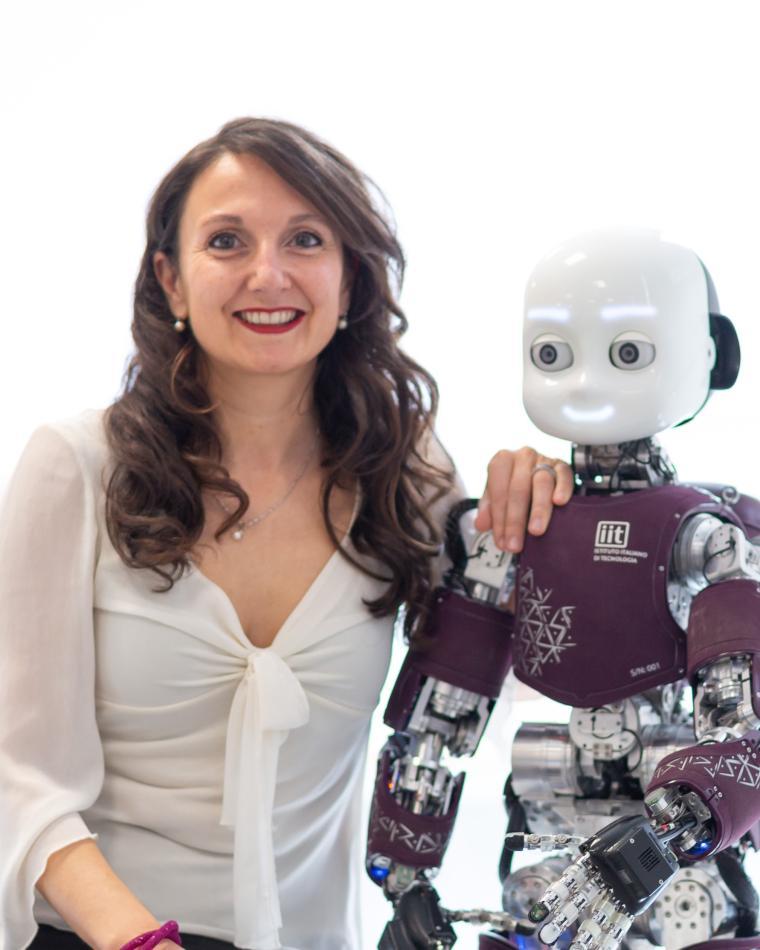}}] {Chiara Bartolozzi} received the Engineering degree (with honors) from the University of Genova, Italy, and the PhD degree in neuroinformatics from ETH Zurich, developing analog subthreshold circuits for emulating biophysical neuronal properties onto silicon and modelling selective attention on hierarchical multi-chip systems. She is currently a researcher with the Istituto Italiano di Tecnologia. She is currently leading the Event-Driven Perception for Robotics Group (www.edpr.iit.it), mainly working on the application of the neuromorphic, engineering approach to the design of sensors and algorithms for robotic perception. She was the chair of the Neuromorphic Systems and Application Technical Committee of IEEE CAS and the co-general chair of IEEE International Conference on AI Circuits and Systems AICAS2020.
\end{IEEEbiography}

\begin{IEEEbiography}[{\includegraphics[width=1in, height=1in, clip, keepaspectratio]{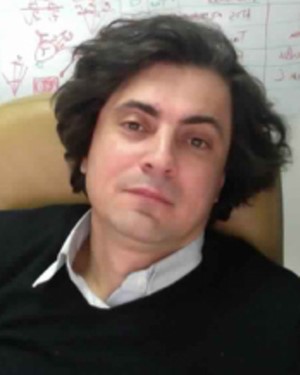}}] {Ryad B. Benosman} received the M.Sc. and Ph.D. degrees in applied mathematics and robotics from the University Pierre and Marie Curie, Paris, France, in 1994 and 1999, respectively.

He is currently a Full Professor with the University of Pittsburgh/Carnegie Mellon/Sorbonne University, Pittsburgh, PA, USA. His work pioneered the field of event-based vision. He is also the Co-Founder of several neuromorphic related companies, including Prophesee, Paris, France—the world leader company in event-based cameras and Pixium Vision, Paris—a retina prosthetics company. He has authored more than 60 publications that are considered foundational to the field of event-based vision. He holds several patents in the area of event vision, robotics, and image sensing. Prof. Benosman was awarded with the National Best French Scientific Article by the Publication LaRecherche for his work on neuromorphic retinas applied to retina prosthetics, in 2013.
\end{IEEEbiography}

\begin{IEEEbiography}[{\includegraphics[width=1.2in, height=1.2in, clip, keepaspectratio]{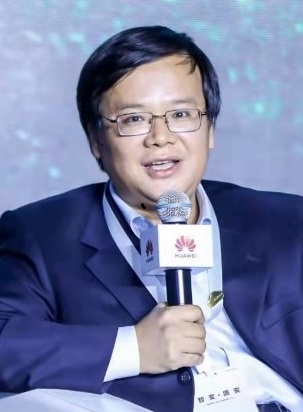}}] {Yonghong Tian \emph{(IEEE Fellow)}} is currently the Dean of the School of Electronics and Computer Engineering, a Boya Distinguished Professor with the School of Computer Science, Peking University, China, and is also the deputy director of Artificial Intelligence Research, PengCheng Laboratory, Shenzhen, China. His research interests include neuromorphic vision, distributed machine learning, and multimedia big data. He is the author or coauthor of over 300 technical articles in refereed journals and conferences. 

Prof. Tian was/is a Senior/Associate Editor of IEEE TCSVT (2018.1-2021.12), IEEE TMM (2014.8-2018.8), IEEE Multimedia Mag. (2018.1-2022.8), and IEEE Access (2017.1-2021.12). He co-initiated IEEE Int’l Conf. on Multimedia Big Data (BigMM) and served as the TPC Co-chair of BigMM 2015, and also served as the Technical Program Co-chair of IEEE ICME 2015, IEEE ISM 2015 and IEEE MIPR 2018/2019, and General Co-chair of IEEE MIPR 2020 and ICME2021. He is a TPC Member of more than ten conferences such as CVPR, ICCV, ACM KDD, AAAI, ACM MM and ECCV. He was the recipient of the Chinese National Science Foundation for Distinguished Young Scholars in 2018 and 2024, two National Science and Technology Awards, and four ministerial-level awards in China, and obtained the 2015 EURASIP Best Paper Award for Journal on Image and Video Processing, and the best paper award of IEEE BigMM 2018, and the 2022 IEEE SA Standards Medallion and SA Emerging Technology Award. He is a Fellow of IEEE.
\end{IEEEbiography}

\vfill

\end{document}